\documentclass[10pt,journal,compsoc]{IEEEtran}
\ifCLASSOPTIONcompsoc
  \usepackage[nocompress]{cite}
\else
  \usepackage{cite}
\fi

\ifCLASSINFOpdf
  \usepackage[pdftex]{graphicx}
  \usepackage{color}
  \usepackage{amsmath}
  \usepackage{amsfonts}
  \usepackage{mathabx}
  
\usepackage{enumitem}   
\else
\fi

\begin{document}

\newcommand\blfootnote[1]{%
  \begingroup
  \renewcommand\thefootnote{}\footnote{#1}%
  \addtocounter{footnote}{-1}%
  \endgroup
}

\newcommand{\red}{\color[rgb]{0.6,0.0,0.0}}
\newcommand{\yellow}{\color[rgb]{0.6,0.6,0.0}}
\newcommand{\green}{\color[rgb]{0.0,0.5,0.0}}
\newcommand{\blue}{\color[rgb]{0.0,0.4,0.6}}
\newcommand{\purple}{\color[rgb]{0.6,0.15,0.7}}

\newcommand{\patrick}[1]{{[\bf \blue {Patrick: } #1}]}
\newcommand{\hc}[1]{{[\bf \green {Henry C: } #1}]}
\newcommand{\ck}[1]{{[\bf \red {Charlie: } #1}]}
\newcommand{\greg}[1]{{[\bf \purple {Greg: } #1}]}

\newcommand{\etal}{\textit{et al.}}

\newcommand{\methodname}{BodyPressure}

%
\title{\methodname~- Inferring Body Pose and \\ Contact Pressure from a Depth Image}

\author{Henry M. Clever*, 
        Patrick Grady, 
        Greg Turk, 
        and Charles C. Kemp 
        
\thanks{Manuscript received April 19, 2005; revised XXXX.}}

\IEEEtitleabstractindextext{%
\begin{abstract}
Contact pressure between the human body and its surroundings 
has important implications. For example, it plays a role in comfort, safety, posture, and health. We present a method that infers contact pressure between a human body and a mattress from a depth image. Specifically, we focus on using a depth image from a downward facing camera to infer pressure on a body at rest in bed occluded by bedding, which is directly applicable to the prevention of pressure injuries in healthcare. Our approach involves augmenting a real dataset with synthetic data generated via a soft-body physics simulation of a human body, a mattress, a pressure sensing mat, and a blanket.  We introduce a novel deep network that we trained on an augmented dataset and evaluated with real data. The network contains an embedded human body mesh model and uses a white-box 
model of depth and pressure image generation. Our network successfully infers body pose, outperforming prior work. It also infers contact pressure across a 3D mesh model of the human body, which is a novel capability, and does so in the presence of occlusion from blankets. 
\end{abstract}


}
\twocolumn[{%
\renewcommand\twocolumn[1][]{#1}%
\maketitle

\begin{center}
\centering
\vspace{-0.7cm}
\includegraphics[width=18.0cm]{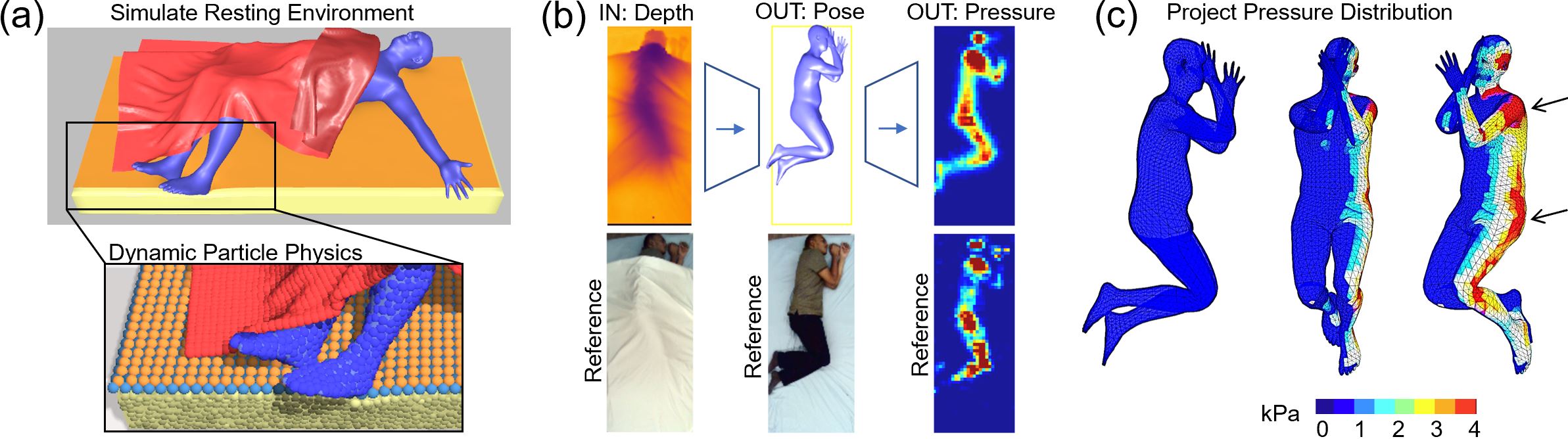}
\end{center}
\vspace{-0.2cm}
\fontfamily{phv} \selectfont{\footnotesize  Fig. 1. We use fast physics simulations to generate \methodname SD, a large synthetic human resting pose dataset, and then train a deep model, \methodname Wnet, to infer pose and pressure from a depth image. (a) Our simulation method rests bodies on a soft bed and covers them with blankets. We then render depth images from the perspective of an overhead camera, and generate pressure images from a pressure mat underneath the person. (b) Using this data, we learn a mapping from depth and gender to pose and contact pressure. (c) This enables a low-cost camera to infer the pressure distribution of the person and potentially detect pressure injuries.}
\vspace{16mm}
}]

\setcounter{figure}{1}  


\IEEEdisplaynontitleabstractindextext

%
\IEEEpeerreviewmaketitle

\IEEEraisesectionheading{\section{Introduction}\label{sec:introduction}} \vspace{-4mm}

%
%
%
%

\blfootnote{\vspace{-0.3cm}\begin{itemize}[leftmargin=3mm]
    \item \textit{ \hspace{-3.2mm} H. M. Clever, P. Grady, and C. C. Kemp are with the Department of Biomedical Engineering, Georgia Institute of Technology, Atlanta, GA, USA. G. Turk is with the School of Interactive Computing, Georgia Institte of Technology, Atlanta, GA, USA. \vspace{2mm} \protect\\
 Email: henryclever@gatech.edu}
    
\vspace{-0.35cm}
\end{itemize}}

\IEEEPARstart{P}{ressure} injuries are an extremely common and longstanding ailment for bedridden individuals, yet technologies to reliably detect them, such as pressure mats, remain expensive and rare in practice. Using a low-cost sensor such as a camera for this task could enable the widespread proliferation of pressure injury detection systems, reducing the 2.5 million of such injuries which occur in the U.S. every year~\cite{arhq2021}. However, sensing pressure from camera imagery faces substantial challenges: not only is the contact interface visually occluded by the human body itself, but the person is frequently covered with blankets, which makes it challenging to even sense where the person is in bed.


We propose a method, \methodname, that can accurately infer body pose and contact pressure from a single image captured by a low-cost depth camera. With these constituents, our method can localize regions of high pressure underneath a person in bed by projecting the pressure onto a human body model. Our approach takes as input a depth image capturing a person underneath blankets, and infers the full 3D human pose and body shape, as well as the contact pressure between the body and the mattress.

To address the challenge of heavy occlusion when inferring human pose at rest, prior work has required multiple modalities as input, including RGB, depth, thermal, and pressure imagery~\cite{liu2020simultaneously, yin2020multimodal}. Our deep network, \methodname Wnet, employs only a depth camera, but when trained on enough data--over $100,000$ samples--it substantially outperforms prior work. The depth modality has a number of benefits: it can be generated easily in simulation by rendering object geometry; deep learning models trained with synthetic depth images transfer well to the real world~\cite{song2017semantic}; and depth imagery preserves patient privacy better than RGB imagery~\cite{luo2018computer}. Further, rather than requiring contact pressure as input during pose estimation, our model \textit{outputs} contact pressure.

To create such a large collection of training samples, we employ fast physics simulations to generate \methodname SD, a synthetic dataset consisting of depth images, body poses, and pressure sensing mat data. We extend our previous work, Clever \etal \cite{clever2020bodies}, which simulates human bodies resting on a soft mattress with a pressure-sensing mat. We extend this method by generating blankets to cover the resting bodies, producing a set of meshes representing the bed, the person in bed, and the blanket covering the person (Fig. 1(a)). We then render these meshes to generate images similar to those captured by a real depth camera. This process can quickly generate data for training data-hungry deep models. We find that the synthetic depth images, body poses, and pressure images resemble their real counterparts with enough fidelity to greatly boost performance of the deep model.


Deep models for human pose estimation are highly sensitive to the pose distribution in the training data. To generate the synthetic bodies at rest, we initialize the simulator with poses that are close to real poses in the Simultaneously-collected Lying Pose (SLP) dataset~\cite{liu2020simultaneously}. As the SLP dataset only has 2D pose annotations, we present an annotation method to fit the 3D Skinned Multi-Person Linear (SMPL) body model~\cite{loper2015smpl} to the real data. This provides high quality pose annotations that are suitable not only for initializing the simulator, 
but are also sufficiently accurate to consider as ground truth for training and testing the deep models. Besides providing a diverse set of real human resting poses in bed, the SLP dataset goes beyond prior datasets in multiple ways. For example, it contains co-registered depth and pressure imagery and the images are recorded with varying scenarios of blanket occlusion, which make it a good candidate for evaluating our methods.

Inferring the pressure distribution underneath a person from a depth image involves three main steps: (1) inferring human pose as a mesh model, (2) inferring the contact pressure on the top surface of the bed, and (3) projecting the pressure from the bed surface onto the body mesh. We introduce a network architecture to address this challenge, which uses a convolutional neural network (CNN) encoder to estimate the parameters for a SMPL parametric human model~\cite{loper2015smpl}. The network contains a differentiable white-box model to reconstruct depth and pressure maps of the estimated human with no learnable parameters. In contrast to many past works that learn to reconstruct images in a black-box neural network~\cite{ronneberger2015u, bulat2016human, rhodin2018unsupervised, oberweger2015training}, the white-box model has a number of advantages, including performance and computation efficiency. 
We use a series of two CNN-based modules to make an initial estimate, and refine it using spatial residual error feedback~\cite{oberweger2015training, carreira2016human, clever2020bodies}.

Although our model inputs a single depth image, it improves pose estimation accuracy by 12\% as compared to the state-of-the-art model~\cite{yin2020multimodal}, which uses RGB, depth, thermal, and pressure imagery as input. Further, we show how it can localize peak pressure areas more accurately than traditional black-box image generation models. 
In summary, our contributions include the following:

\begin{itemize}

\item A method, \methodname\footnote{Code  and data: \texttt{\scriptsize{github.com/Healthcare-Robotics/BodyPressure}}}, that takes as input depth images of a person in bed from an overhead camera and infers the body pose, contact pressure, and localized regions of high pressure density.
\item \methodname SD, a synthetic dataset consisting of $97,495$ bodies at rest with pressure images and depth images rendered with and without simulated blankets.
\item SLP-3Dfits, a dataset consisting of $4,545$ SMPL bodies~\cite{loper2015smpl} fit to the SLP dataset~\cite{liu2020simultaneously} that resolves depth perspective ambiguity by fitting to the point clouds from the dataset.

\end{itemize}

Section~\ref{sec:lit_review} covers related literature. Section~\ref{sec:slp_optim} presents our real data annotation method; we use the 3D human body annotations in the following sections. Section~\ref{sec:methods1} presents our synthetic data generation pipeline with physics simulation. Section~\ref{sec:methods2} presents our deep network architectures, which are trained using real and synthetic data. Section~\ref{sec:evaluation} explains how we evaluate our method, followed by the results and discussion in Section~\ref{sec:results}.

\section{Related work}\label{sec:lit_review}


\begin{figure*}
\begin{center}
\includegraphics[width=17.5cm]{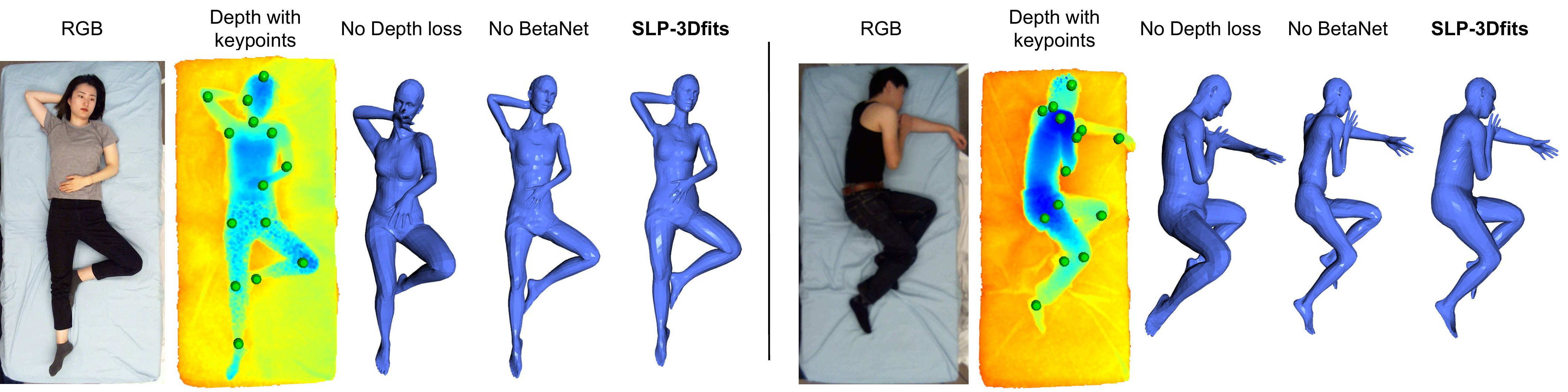}
\end{center}
\vspace{-0.5cm}
\caption{We fit 3D SMPL bodies to the SLP dataset~\cite{liu2020simultaneously}, which we use for initializing the physics simulator and for training and testing our deep models. Our method resolves depth ambiguity using a loss between the SMPL mesh and 3D points from the depth image. Examples are shown without the depth loss term, resulting in poses with depth error. Examples are also shown without BetaNet, resulting in bodies with unreasonable shapes.}
\label{fig:slp_fitting}
\end{figure*}

\textbf{Sensor-based pressure injury monitoring.} Commercial pressure mapping systems are among the most common methods of monitoring pressure injury risk, and have been used to more effectively reposition patients and reduce high pressure areas~\cite{mansfield2019pressure}. Researchers have made progress to improve monitoring through automatic bodypart localization~\cite{farshbaf2013detecting, liu2014bodypart, pouyan2016automatic} and posture detection~\cite{yousefi2011bed}. An alternative to this is wearable pressure sensors that can adhere to at risk areas~\cite{mansfield2019objective}. The cost of these devices can deter widespread use, so others have studied how inertial measurement units (IMUs) can be used~\cite{pickham2018effect}, among other methods~\cite{mansfield2019pressure}. Yet, peak pressure localization remains a challenge. The sacrum and heels have been noted as the most common areas, but also occur on the hips, elbows, ischium, shoulders, spinous process, ankles, toes, and head~\cite{vanderwee2007pressure, lahmann2006pressure}.


\textbf{Humans at rest.}
While many works in computer vision model humans in \textit{active} poses such as pedestrians crossing the street~\cite{ghori2018learning}, \textit{resting} belongs to a different class of human activity. 
Resting is characterized by a low degree of physical exertion, substantial contact with surrounding surfaces such as a bed or chair, and the fact that people spend an overwhelming portion of life resting. With the ability to learn complex mappings between images and labels using CNNs, researchers have inferred human resting pose from diverse human configurations, postures, and sensing modalities~\cite{achilles2016patient, casas2019patient, clever20183d, chen2018patient, liu2019deep, liu2019seeing, liu2020simultaneously}. 

Maintaining awareness of scene constraints and dynamics can enable more physically plausible models of humans at rest. Chao \etal~\cite{chao2019learning} used reinforcement learning to teach dynamic agents how to sit on a chair in a virtual environment. Hassan \etal~\cite{hassan2019resolving} used optimization to infer pose in a way that the human model is consistent with its surroundings, i.e. not floating above a chair or sunk into it unrealistically. Our previous work modeled humans in bed using ragdoll physics~\cite{clever2020bodies}, and another work synthesized human poses in arbitrary environments with objects that could be contacted or rested upon~\cite{zhang2020generating}.

\textbf{Simulating human environments.}
Approaches for generating synthetic data that model humans in the context of deep learning use physics simulators such as DART \cite{lee2018dart} and PyBullet \cite{tan2018sim, erickson2019assistivegym} and position-based dynamics simulators such as PhysX \cite{erickson2018deep} and FleX \cite{macklin2014unified}. In a recent work, we combined DART and FleX to rest kinematic human bodies on a soft mattress~\cite{clever2020bodies}, and randomized the human pose and body shape to increase variability. Others have explored cloth with physics simulations \cite{clegg2017learning, cusumano2011bringing,  erickson2018deep, matas2018sim}, which could be used to create a diverse set of blanket configurations and profiles on a person resting in bed.

\textbf{Simulating pressure and depth images.} 
We refer the reader to our previous work on simulating pressure imagery~\cite{clever2020bodies}, which includes a pressure image generation method that we use. For vision, RGB image synthesis relies on relatively complex graphics approaches \cite{chen2016synth, varol2017learning, yu2019simulcap} while creating synthetic depth images is more straightforward~\cite{shotton2011real,martinez2018real}. Achilles \etal \cite{achilles2016patient} generated depth data for a bed environment by simulating a blanket covering the person, and trained a deep network using this data. While this work is close in concept to ours, the human is represented with a skeleton, which has limitations, and the code and dataset are unavailable. 
To improve generalization performance, researchers have used noise models with pixel dropout, spot noise, and synthetic occlusion~\cite{liu2020simultaneously, shotton2011real, zhong2017random, planche2017depthsynth}. Others have denoised real data during test time~\cite{zakharov2018keep}, at the cost of real-time inference speed.


\textbf{Annotating datasets with 3D human mesh models.} Standard human pose datasets contain 2D keypoint annotations, which are pixel-wise joint position coordinate labels on images. Researchers have fit 3D human mesh models to these 2D keypoints, by projecting a 3D body into image coordinates and optimizing over the human model parameters (e.g. kinematic joint angles) ~\cite{bogo2016keep, kolotouros2019learning}. Yin \etal~\cite{yin2020multimodal} use the SPIN method~\cite{kolotouros2019learning} for fitting SMPL bodies to the SLP resting pose dataset. However, these methods suffer from depth perspective ambiguity. When additional information is present, such as 3D point clouds or scene geometry, it is possible to resolve these ambiguities, which Hassan \etal~\cite{hassan2019resolving} showed. This can provide highly accurate annotations, but such optimizations require careful data preprocessing and are too slow to use during inference time. 

\textbf{Deep learning for 3D human pose estimation.} Inferring 3D human pose is a significant branch of research in computer vision~\cite{yin2020multimodal, clever2020bodies, rhodin2018unsupervised, hassan2019resolving, chen2016synth,  varol2017learning, bogo2016keep, kanazawa2018end,  kolotouros2019learning, pavlakos2017coarse}. In recent years, deep learning has seen widespread use for inferring human pose, by using convolutional neural networks (CNNs) to encode features in an image and output some representation of human pose. We refer the reader to literature surveys for more comprehensive coverage \cite{sarafianos20163dhuman, zheng2020deep}. Here we discuss approaches that are relevant to the particular black-box and white-box architectures we use. Differentiable kinematic models embedded into image encoders have gained traction in research due to their ability to produce physically plausible human models~\cite{zhou2016deep, clever20183d, kanazawa2018end, hasson2019learning, Grady_2021_CVPR}. In contrast to differentiable skeleton models, parametric human mesh models such as SMPL~\cite{loper2015smpl} offer a better representation of body shape and size. Both of our architectures use this. 

Many pose estimation methods incorporate black-box image reconstruction for purposes including heatmap regression~\cite{zhou2018monocap}, geometry awareness~\cite{rhodin2018unsupervised}, and spatial residual error correction~\cite{oberweger2015training}. One such architecture that may be used for this is U-Net~\cite{ronneberger2015u}, which learns an image-to-image mapping with a latent space in the middle that can encode image classes~\cite{harouni2018universal} or physically meaningful features~\cite{he2018learning}. Fewer deep learning works have used white-box methods for image reconstruction, i.e. differentiable image generation methods with no learnable parameters. However, our previous work introduced a model of pressure map reconstruction~\cite{clever2020bodies} and was trained only with synthetic data, which we build on.

\section{Annotating Real Data with SMPL Bodies}\label{sec:slp_optim}

Here we describe an optimization method for annotating an existing human pose dataset with 3D SMPL bodies, as shown in Fig.~\ref{fig:slp_fitting}. This method finds body shape and pose parameters to fit the SMPL bodies to depth images and existing 2D keypoint annotations. The optimization includes terms for scene constraints, body mass, and height, if they are available in the dataset. By leveraging depth information, the method can resolve the ambiguities in pose which are inherent with 2D keypoint annotations alone. The method can produce high-quality fits even for complex poses. We annotate the SLP dataset~\cite{liu2020simultaneously} to create SLP-3Dfits, a dataset consisting of 3D fits to $4,545$ unique poses as described in Section~\ref{ssec:eval_vp}. 
SLP-3Dfits is used for initializing the synthetic data generation method described in Section~\ref{sec:methods1}, and evaluating the deep learning methods described in Section~\ref{sec:methods2}. 

Our optimization requires a depth image, $\mathcal{D}$, capturing a real person who is not occluded by objects in the environment (e.g. blankets or coverings). The depth image is projected into 3D point cloud space; the SMPL mesh is initialized in this 3D space; and the optimization begins. A loss is computed between $K$ 2D keypoints $\boldsymbol{\text{S}} \in \mathbb{R}^{K \times 2}$ and 3D SMPL joint positions projected into image space. Scene constraints, $\mathbb{C}$, are used to reduce interpenetration between the body and the bed. Height and body mass measurements, $h$, $m$, are used to enforce physical consistency in the SMPL body shape, by modeling height and body mass as a function of SMPL body shape parameters. This is described in Section~\ref{ssec:betanet}. We define the SMPL body parameters with  $\boldsymbol{\Psi}_R = \big[ \boldsymbol{\beta}_R \hspace{3mm} \boldsymbol{\Theta}_R \hspace{3mm} \boldsymbol{s}_R \hspace{3mm} \boldsymbol{\phi}_R \big]^{\top}$, which contains the body shape parameters $\boldsymbol{\beta}_R$, the joint angles $\boldsymbol{\Theta}_R$, and the global translation and rotation $\boldsymbol{s}_R$, $\boldsymbol{\phi}_R$. Subscript $R$ distinguishes the real data annotations from parameters in later sections. The optimization seeks $\boldsymbol{\Psi}_R$ that minimizes the objective function $E$ as follows:


\begin{multline}
E(\boldsymbol{\Psi}_R, \mathcal{D},\boldsymbol{\text{S}}, \mathbb{C}, h, m) = E_J + \lambda_{\mathcal{D}} E_{\mathcal{D}} + \lambda_{P}E_P \\ + \lambda_{M}E_M + \lambda_{\beta}E_{\beta}
\label{eqn:slp_fits_error_fn}
\end{multline}

The error function contains the following terms:

\begin{itemize}
    \item $E_J(\boldsymbol{\Psi}_R, \boldsymbol{\text{S}})$ penalizes the distance between 2D keypoint annotations and SMPL joints projected into camera space.
    \item $E_{\mathcal{D}}(\boldsymbol{\Psi}_R, \mathcal{D})$ encourages a match between depth points and SMPL vertices visible from the camera's perspective. This is implemented using a robust version \cite{hassan2019resolving} of the Chamfer Distance \cite{fan2017point}.
    \item $E_{M}(\boldsymbol{\Psi}_R, \mathbb{C})$ enforces scene constraints by penalizing interpenetration between the body and the scene (bed, mattress).
    \item $E_{P}(\boldsymbol{\Psi}_R)$ penalizes body self-penetration, e.g., a hand interpenetrating through the chest. Collisions are detected using Bounding Volume Hierarchies \cite{ballan2012motion}.
    \item $E_{\beta}(\boldsymbol{\Psi}_R, h, m)$ encourages the SMPL shape parameters $\boldsymbol{\beta}_R$ to express a body with a height and mass that match the participant's measurements $h, m$. This requires a mapping from body shape to body height and mass, which is described in Section \ref{ssec:betanet}.
\end{itemize}

During optimization, the SMPL joint angles $\boldsymbol{\Theta}_R$ are constrained to nominal textbook values for joint angle limits~\cite{boone1979normal, roaas1982normal, soucie2011range}, reducing the probability of generating contorted poses. Further, many datasets such as the SLP dataset~\cite{liu2020simultaneously} contain a variety of different poses for a particular person. Accordingly, all such posed bodies should have the same shape. To ensure that the SMPL shape parameters for each participant are identical across all dataset samples, the optimization was implemented as a batched optimization, allowing the joint optimization of body pose and shape across multiple samples. 

As the optimization of SMPL parameters $\boldsymbol{\Psi}_R$ provides a non-convex objective, the optimization may fall into local minima. To avoid this, each sample is initialized and optimized multiple times with different starting poses and orientations. The result with the lowest loss is selected.

The optimization problem is solved using the ADAM differentiable optimizer. The method is similar to the approach used by Hassan \etal \cite{hassan2019resolving}; however, notable additions include our method of enforcing physical consistency on body height and mass, and our batched optimization that fits the same body shape across multiple pose samples.


\begin{figure*}
\begin{center}
\includegraphics[width=17.5cm]{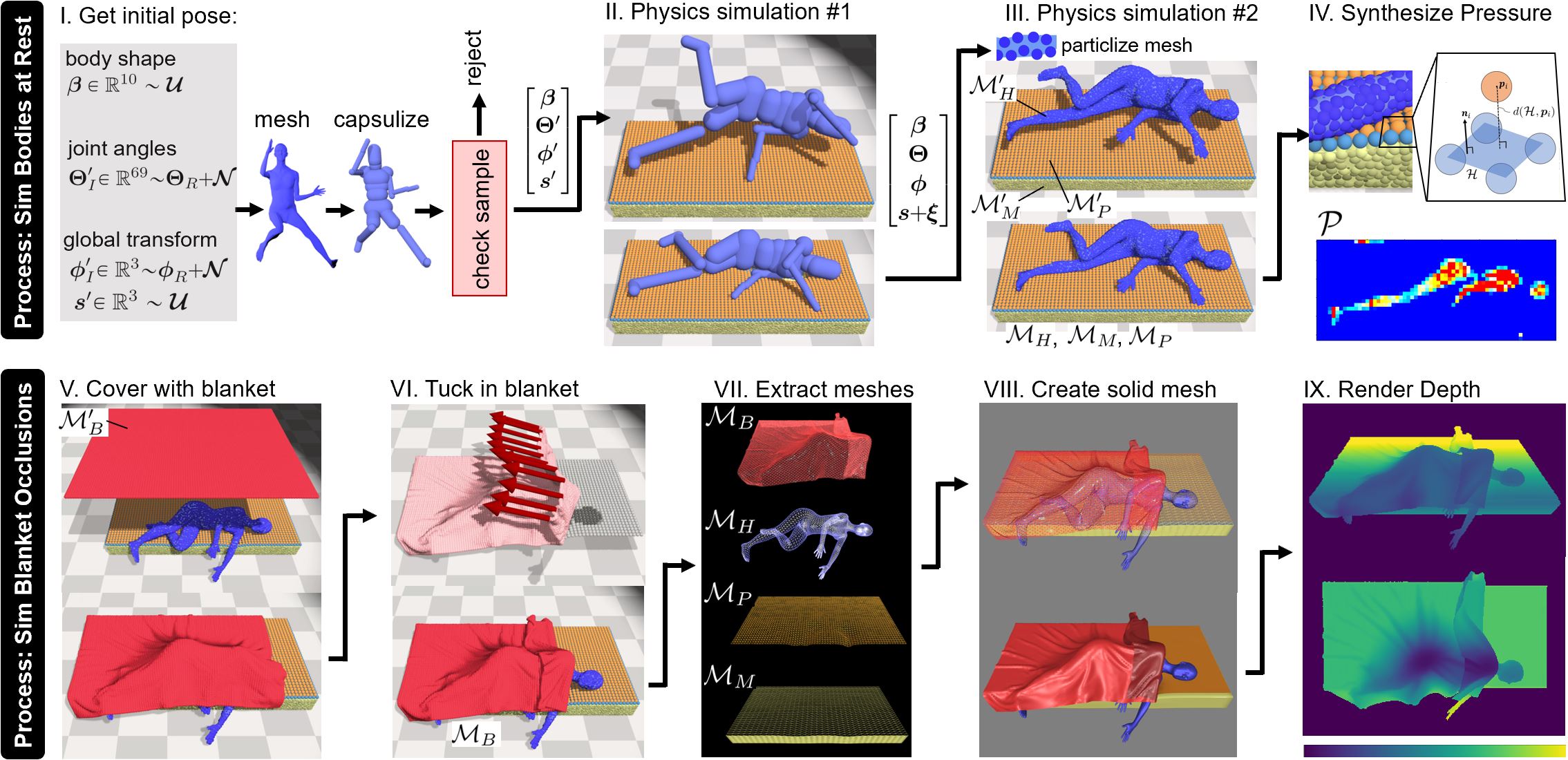}
\end{center}
\vspace{-0.5cm}
\caption{Our synthetic data generation method involves two processes: The first process is similar to that of our previous work~\cite{clever2020bodies}; it involves (I.) sampling random human poses, (II.) resting dynamic capsulized bodies on a soft bed to find a resting pose, (III.) resting a finer body representation on the bed to improve human body shape detail, and (IV.) using a simulated pressure sensing mat underneath the person to compute a pressure image. The second process involves (V.) covering the body with a blanket, (VI.) pulling the top of the blanket down to uncover the person's head, (VII.) extracting deformed meshes, (VIII.) creating a solid mesh, and (IX.) simulating a depth imagery from a pinhole camera positioned above the bed.}
\label{fig:pipeline}
\end{figure*}

\subsection{BetaNet}\label{ssec:betanet}
To calculate $E_{\beta}$, we model body height and mass as a function of SMPL body shape parameters and gender, i.e. $\{h, m\} = f_{\beta}(\boldsymbol{\beta}, \boldsymbol{g})$, where $h$ and $m$ are values in units of meters and kilograms, respectively. Unlike $\boldsymbol{\beta}$, height and weight are directly measurable physical values that can better constrain the network. Gender is modeled with 
two flags, i.e. $\boldsymbol{g}\in \mathbb{R}^2$, which may account for female ($[0,1]$), male ($[1,0]$), or gender-neutral ($[1,1]$) body models. In this work however, only female and male models are used.  
We represent the function $f_{\beta}$ with a 2-layer fully connected network, where the input consists of body shape parameters and gender, and the output is height and body mass. We train BetaNet on a large synthetic dataset consisting of randomly shaped SMPL bodies with known height and mass, and mass is modeled as a function of SMPL body mesh volume:

\begin{equation}{
m = \bar{m}_{\boldsymbol{g}} {\mathcal{V}_{mesh, \boldsymbol{g}} \over \bar{\mathcal{V}}_{mesh,\boldsymbol{g}} }}
\label{eqn:m_vol_ratio}
\end{equation}

where $\bar{m}_{\boldsymbol{g}}$ is a gender-specific average body mass value we take from Tozeren~\cite{tozeren1999human}, $\mathcal{V}_{mesh, \boldsymbol{g}}$ is the volume of a gendered body model of interest, and $\bar{\mathcal{V}}_{mesh, \boldsymbol{g}}$ is the volume of a gendered body model with average shape $\boldsymbol{\beta} = \boldsymbol{0}$. We train BetaNet with the following loss function:

\begin{equation}
\mathcal{L}_{\textrm{BetaNet}} := {1\over \sigma_{h}} ||h - \hat{h}||_1 + {1\over \sigma_{m}} ||m - \hat{m}||_1 
\label{eqn:betanet}
\end{equation}

This is normalized by standard deviations $\sigma_{h}$ and $\sigma_{m}$, which are computed from the entire synthetic training dataset. The trained BetaNet is also used for the separate problem of learning a mapping from depth to pose and pressure, described in Section~\ref{sec:methods2}.

\subsection{SLP-3Dfits: SMPL Fits for 4545 Real Poses}\label{ssec:eval_vp}

Here we describe how our method of fitting parametric SMPL bodies to depth images and keypoints is applied to the SLP~\cite{liu2020simultaneously} dataset, which contains $4,590$ poses across $102$ participants. We use these fits for generating synthetic data, training our deep network, and testing it. 

We annotate $4,545$ poses across $101$ participants (one subject is excluded due to a calibration issue). 
The SLP dataset contains calibrated, occlusion-free depth images (the `uncovered' case) for every pose. `Calibrated' in this context means that the depth images are spatially co-registered to other modalities and annotations using the calibration method in Liu \etal~\cite{liu2020simultaneously}. After converting these to point clouds, points from the bed surface are filtered out, so that all points used in the objective function are from the surface of the human body. The SLP dataset also contains 2D keypoint annotations, body height, and body mass for all captured poses, which are factored into the objective function. A plane representing the height of the bed surface is used as a scene constraint, so the body and limbs do not penetrate the mattress too far.

Generally the fits from the automatic optimization are of high quality. However, in some cases the result converges to an incorrect local minimum, usually when the participant is lying on their side and the hands are posed on the wrong side of the head. Thus, each result of the fitting process is manually checked by a human annotator for agreement with the original pose in the image data. For failure cases, the optimization is restarted with a different initialization and then re-checked. Roughly $9 \%$ of the fits required these restarts. The fits are of sufficient accuracy to use as "ground-truth" for evaluation of neural network predictions, and we have made them available publicly. The mean error between the SMPL joint locations and the 2D skeleton annotations is 41.6 mm; per-joint error is provided in Appendix~\ref{sec:app_3dfits}. Note that the SLP dataset joint annotations have some offset when compared to the SMPL model joints, inflating this error metric.

\section{Synthetic Data Generation}\label{sec:methods1}

We present a synthetic data generation pipeline using physics simulations that is capable of creating a large dataset of humans resting. It can generate bodies at rest on a soft mattress with depth and pressure images. In practice, this approach is much more efficient than collecting comparable real-world data. The sole purpose of this pipeline is to create a large dataset, \methodname SD, for training the deep model described in Section~\ref{sec:methods2}. \methodname SD contains $97,495$ unique body shapes, poses, and image samples; data partitions are described in the evaluation (Section~\ref{ssec:bodypressure_sd}).

\begin{figure*}
\begin{center}
\includegraphics[width=17.5cm]{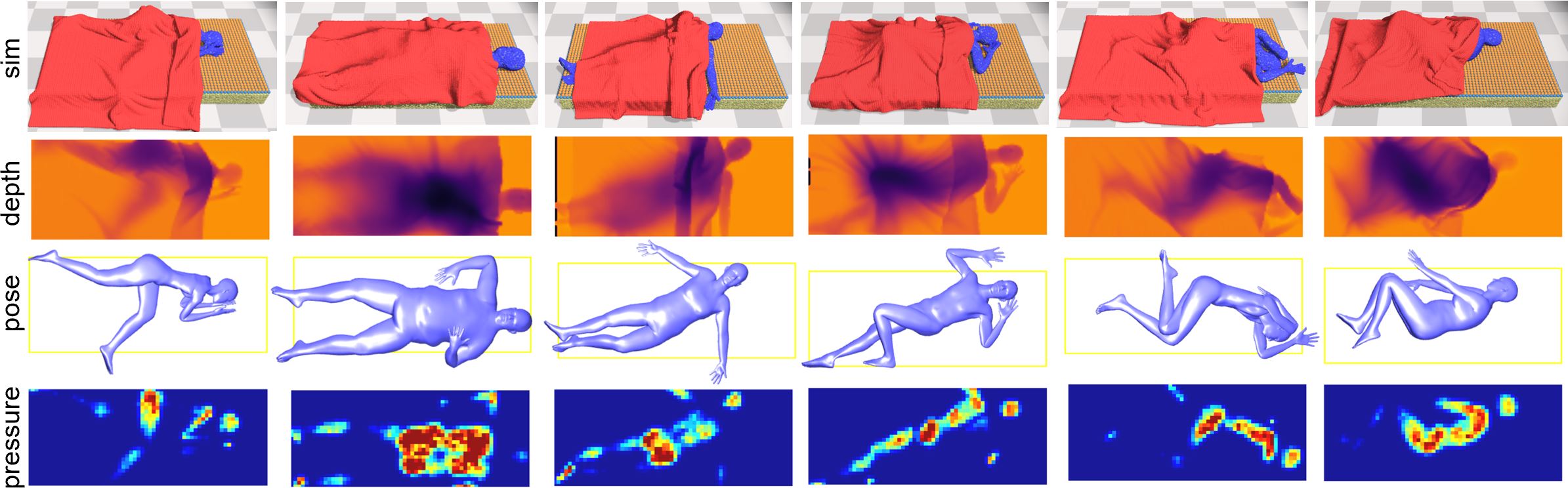}
\end{center}
\vspace{-0.5cm}
\caption{\methodname SD synthetic data samples created by resting bodies on a mattress and covering them with a blanket.}
\label{fig:dataset_examples}
\end{figure*}

The data generation pipeline consists of two processes, as depicted in Fig.~\ref{fig:pipeline}. The first process (I. - VI.) is similar to that of our previous work~\cite{clever2020bodies}; it generates bodies resting on a soft bed with synthetic pressure images. The second process (V. - IX.) generates blanket occlusions on top of the bodies in bed with synthetic depth images. It uses three simulation tools: DART~\cite{lee2018dart} for simulating articulated human dynamics;  FleX~\cite{macklin2014unified} for simulating soft materials that include the human body, bed mattress, pressure sensing mat, and blanket dynamics; and PyRender~\cite{matl2020pyrender} for depth image rendering. Some synthetic data examples are shown in Fig.~\ref{fig:dataset_examples}. 

\subsection{Simulating Bodies at Rest} \label{ssec:bodies}

The process begins by sampling a large set of initial synthetic body poses, where each pose sample contains joint angles $\{ \boldsymbol{\Theta}_I',  \boldsymbol{\phi}_I' \}$, that are close to the real poses $\{ \boldsymbol{\Theta}_R,  \boldsymbol{\phi}_R \}$ fit in Section~\ref{sec:slp_optim} This is depicted in Fig.~\ref{fig:pipeline}-I. Normally distributed noise is added to the hip, knee, inner shoulder, outer shoulder, and elbow joints. Each angle $j$ in these joints receives noise with the following equation: 

\begin{equation}
    \theta_{I,j}' = \theta_{R,j} + \mathcal{N}(0, \pi / 12 )
\end{equation}

where $\{\theta_{I,1}',\theta_{I,2}',... \}= \boldsymbol{\Theta}_I'$ and $\{\theta_{R,1}, \theta_{R,2},... \} = \boldsymbol{\Theta}_R$. Other joints 
are set equal to the real fit angles. The same amount of noise is added to two of the root angle joints representing the rotation of the body along its longitudinal axis ($\phi_{R,2}$) and the rotation of the body normal to gravity ($\phi_{R,3}$). 
No noise is added to $\phi_{R,1}$, which represents rotation around the sagittal axis. For each pose, the body shape is sampled from a uniform distribution, following~\cite{ranjan2020learning}: $\boldsymbol{\beta} \sim \mathcal{U}[ -3, 3]$. The 2D translation of the human body over the surface of the bed is also sampled from a uniform distribution: $s_{1}', s_{2}' \sim \mathcal{U}[-0.2, 0.2]$. The height of the body normal to gravity, $s_{3}'$, is set according to the lowest initial point on the body so that every part of the body is initially above the bed. Accordingly, $\{ s_{1}', s_{2}', s_{3}' \} = \boldsymbol{s}'$. With the fully parameterized body of initial shape and pose $\{\boldsymbol{\beta}, \boldsymbol{\Theta}_I',  \boldsymbol{\phi}_I', \boldsymbol{s}'\}$, body self-collisions are checked using the capsulized body from SMLPify~\cite{bogo2016keep}. 
If there is a collision, the sample is rejected, otherwise, the set of parameters becomes $\{\boldsymbol{\beta}, \boldsymbol{\Theta}',  \boldsymbol{\phi}', \boldsymbol{s}'\}$. 


\textbf{Two physics simulations.} The first process uses physics simulations to convert a body with the initial collision-free pose $\{\boldsymbol{\beta}, \boldsymbol{\Theta}',  \boldsymbol{\phi}', \boldsymbol{s}'\}$ to a resting pose $\{\boldsymbol{\beta}, \boldsymbol{\Theta},  \boldsymbol{\phi}, \boldsymbol{s}\}$ with human body mesh $\mathcal{M}_H$ and pressure image $\mathcal{P}$. It is the same simulation process as our previous work~\cite{clever2020bodies}, with the exception of the weighting method for the simulated body, which we updated. See details in Appendix~\ref{sec:app_capsule_weight}. The remainder of Section~\ref{ssec:bodies} provides a high-level summary of the methods from \cite{clever2020bodies}.

In Physics Simulation \#1 (Fig.~\ref{fig:pipeline}-II.), 
the human is modeled as an articulated rigid body made with capsule primitives. DART~\cite{lee2018dart} is used to model this capsular human and simulate its dynamics. The articulated body uses the same joint angles and body shape parameters as the SMPL mesh, but unlike the mesh, the joint angles can change due to applied torques and forces (e.g. due to gravity and contact with the bed). At the same time, a different simulator, FleX~\cite{macklin2014unified}, is used to model a mattress and a pressure-sensing mat underneath the body. FleX uses a unified particle representation to efficiently model deformable objects. These are combined in a loop to allow a dynamic articulated system (i.e., the body) to interact with soft materials (i.e., the pressure sensing mat).


While the capsulized articulated rigid body from Physics Simulation \#1 is well suited for modeling ragdoll physics and finding a resting pose, it does not represent the surface geometry of the human body with sufficient fidelity for pressure image generation. The process assigns the resting pose and body shape to a SMPL mesh and fills it with deformable FleX particles, which creates a non-articulated body with a finer profile of human features. This is depicted as Physics Simulation \#2 in Fig.~\ref{fig:pipeline}-III. This `particlized' body is positioned above the bed with parameters $\{\boldsymbol{\beta}, \boldsymbol{\Theta},  \boldsymbol{\phi}, \boldsymbol{s} + \boldsymbol{\xi}\}$, where the term $\boldsymbol{\xi}$ represents a vertical adjustment in the root joint translation so the body can be allowed to fall a short distance to settle on the bed a second time. 

The process uses polygon meshes to record the simulation state in Physics Simulation \#2. Initially, the meshes include the undeformed human body, mattress, and pressure sensing mat meshes. The human body mesh is a function of the resting pose: $\mathcal{M}_H' = f\big(\boldsymbol{\beta}, \boldsymbol{\Theta},  \boldsymbol{\phi}, \boldsymbol{s} + \boldsymbol{\xi}\big)$. 
The mattress underneath the person is set to a twin size, consisting of a rectangular prism of particles in an undeformed mesh $\mathcal{M}_M'$. 
The mattress is constructed using the same particlizing method as the human body~\cite{clever2020bodies}. The padded mat on the bed surface represents a layer of bedding underneath the person, and is constructed from a two-layer lattice of particles laced together by a grid of springs constraints, represented by mesh $\mathcal{M}_P'$. 
Physics Simulation \#2 is run until the particlized human body reaches static equilibrium, and then outputs mesh data for the resting human body $\mathcal{M}_H$, the deformed mattress $\mathcal{M}_M$, and the deformed pressure sensing mat $\mathcal{M}_P$.

\subsubsection{Synthesizing Pressure Imagery}
Besides interacting physically with the capsulized body in Physics Simulation \#1, the pressure sensing mat on the surface of the bed is also used to generate pressure images, based on particle penetration between the top (orange) and bottom (blue) layers of particles on the surface of the bed, shown in Fig.~\ref{fig:pipeline}-IV. The simulated sensor measures pressure as a function of 
how far a top layer particle penetrates the underlying particles. Each penetration distance across the mat is converted into a value on pressure image $\mathcal{P}$, which is also shown in Fig.~\ref{fig:pipeline}-IV.

\subsection{Simulating Blanket Occlusions}\label{ssec:blankets}

Before simulating the cloth blanket, the process freezes the ending equilibrium state of Physics Simulation \#2, which involves freezing particles representing the resting human, deformed mattress, and deformed pressure sensing mat. A blanket is created in FleX with a grid of particles shown in Fig.~\ref{fig:pipeline}-V. The blanket is parameterized by two sets of terms: the blanket geometry terms, which determine blanket size, particle location, particle connection points and initial world transform; and the dynamic simulation terms, which determine the stiffness holding particles together.

\begin{figure*}
\begin{center}
\centering
\includegraphics[width=18.2cm]{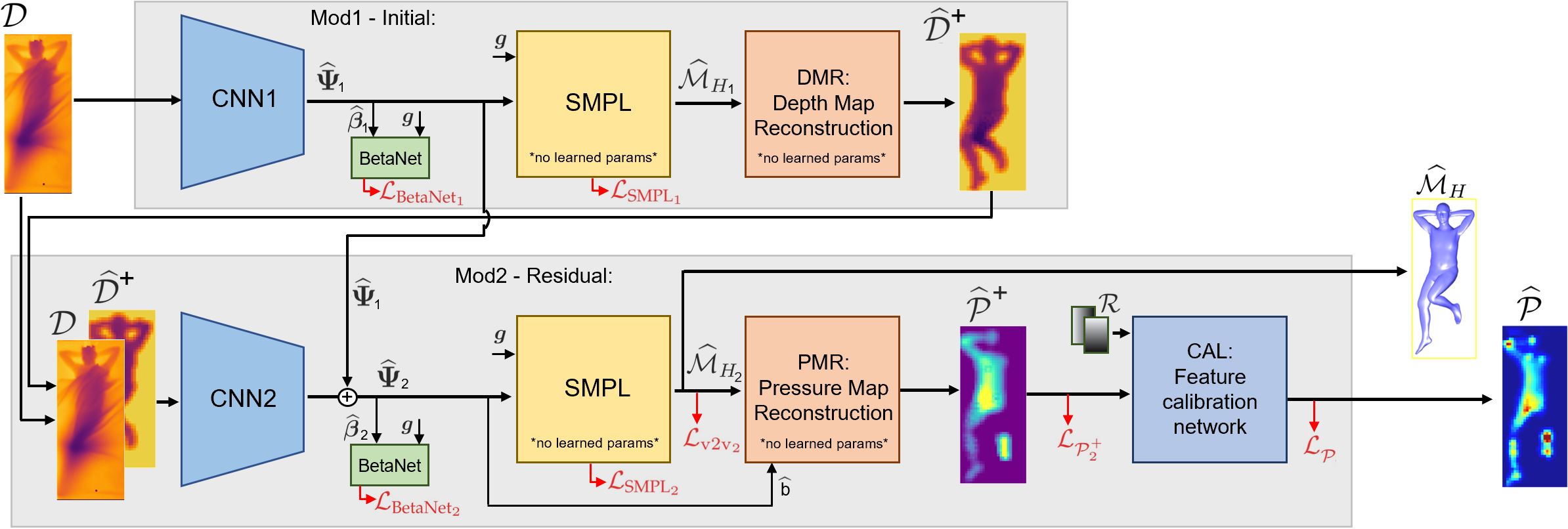}
\vspace{-9mm}
\end{center}
\caption{\methodname Wnet (BPWnet), a deep network that learns a mapping from depth and gender to pose and contact pressure. Depth, $\mathcal{D}$ is encoded with a black-box CNN and outputs SMPL~\cite{loper2015smpl} human model parameters $\boldsymbol{\widehat{\Psi}}$, which is used to reconstruct a SMPL human mesh $\mathcal{\widehat{M}}_H$. Using white-box image reconstruction components DMR and PMR, it refines the pose estimate and outputs a pressure map. The pressure map features are calibrated with CAL to produce a contact pressure estimate $\mathcal{\widehat{P}}$. The two module design refine estimates with initial and residual stages; subscripts 1 and 2 indicate estimates from each module, respectively.
}
\label{fig:networks}
\end{figure*}

The blanket has an undeformed height and width $h_B$ and $w_B$, which are chosen to represent a twin size of $1.68 \times 2.29$ meters, 
and are created with a $102 \times 102$ particle grid. The global translation is parameterized by $\{s_{B,1}, s_{B,2}, s_{B,3}\} = \boldsymbol{s}_B \in \mathbb{R}_3$ and rotation by $\{\phi_{B,1}, \phi_{B,2}, \phi_{B,3}\} = \boldsymbol{\phi}_B \in \mathbb{R}_3$. These can be used to incorporate domain randomization (e.g. by sampling random initial blanket translations) or to better match blanket configurations in a particular dataset. The blanket is initially set to a height $s_{B,3}$ above the body so that the blanket does not initially collide with the body. The initial blanket configuration may be described by mesh $\mathcal{M}_B' = f( h_B, w_B, \boldsymbol{s}_B, \boldsymbol{\phi}_B)$. The blanket is weighted with the same density as the mattress in the previous simulation. 
The blanket dynamics are also influenced by the blanket stiffness $K_B$.

The process then runs the simulation, dropping the soft blanket on the body in bed. Depending on the initial position over the surface of the body, the blanket may cover the human's head, which is undesirable because it would likely not be a common occurrence in the real world. Thus, the blanket is adjusted by pulling on a set of particles on the top edge of the blanket - see Fig.~\ref{fig:pipeline}-VI. 
To determine if the blanket should be pulled to uncover the person's head, the process checks if the blanket is above the person's neckline.  
In other cases, the top edge of the blanket may be initialized at a location very far from the person's head, which could lead to the person being only partially covered. In this case, the algorithm checks if the blanket is below the person's neckline, and if it is, the same set of particles is pulled upward to better cover the person. 
Once the blanket reaches static equilibrium, the simulator halts and outputs the deformed blanket as mesh $\mathcal{M}_B$.

\subsubsection{Rendering Synthetic Depth Imagery} 
The process extracts the deformed meshes $\{\mathcal{M}_H, \mathcal{M}_M, \mathcal{M}_P, \mathcal{M}_B\}$ as depicted in Fig.~\ref{fig:pipeline}-VII., and records them as part of the dataset. Then, they are assembled using Pyrender~\cite{matl2020pyrender}, an open source python library for rendering and visualization (Fig.~\ref{fig:pipeline}-VIII). 
Finally, the process renders $\mathcal{D}$, a depth image from a camera facing the bed. Fig.~\ref{fig:pipeline}-IX depicts two viewing angle perspectives, which include an observation from the side of the bed and an observation from mounting the camera directly above the bed facing downward.

\section{Learning Pose and Pressure from Depth}\label{sec:methods2} 

We train an algorithm that learns a mapping $f$ from a depth image $\mathcal{D}$ captured over a person at rest with gender $\boldsymbol{g}$, to a human mesh $\mathcal{M}_H$ that models pose and body shape, and a 2D array $\mathcal{P}$ that encodes the contact pressure on the surface of the mattress underneath the person:

\begin{equation}
\{\mathcal{M}_H, \mathcal{P}\} = f(\mathcal{D}, \boldsymbol{g})
\label{eqn:Pb_fD}
\end{equation}

We represent $f$ in the form of a deep network.  
The body mesh $\mathcal{M}_H$ and the pressure array $\mathcal{P}$ can be used to calculate the pressure distribution on the surface of the body, $\mathcal{P}_b$ (recall Fig. 1 (c)), which contains localized pressure on specific body parts. We define $\mathcal{P}_b$ as a collection of human body mesh vertices that are each assigned a pressure value due to contact from the underlying surface. Because the mesh $\mathcal{M}_H$ is learned with a 6-DOF pose in the world reference frame and the contact pressure array $\mathcal{P}$ is collected by a sensor mounted at a known position in the world, they are implicitly co-registered. Assuming that the pressure mat exists within a flat plane normal to gravity on the top surface of the bed, 
each contact pressure element $p_{xy}$ may be projected normal to gravity onto the human mesh, where $p_{xy}$ is assigned to mesh vertex $v_j$ if its taxel area contains the $x,y$ position of $v_j$. This mapping approximates the complex phenomena that occur when the mat is deformed due to contact by neglecting stretching and folding. As such,

\begin{equation}
\mathcal{P}_b : \mathcal{P}  \mapsto \mathcal{M}_H
\label{eqn:proj_M}
\end{equation}

This projection is sufficient to localize pressure on specific body parts, because vertex indices on the SMPL model are independent of body pose and shape, i.e. a heel vertex is always on the heel. Vertices with non-zero pressure are required to be unoccluded from the pressure mat by other body parts or surfaces of the body. This is ensured by casting a ray downwards from each vertex toward the mat and checking if it passes through any triangular faces. If it passes through a face, the pressure is set to zero. This will ignore pressure due to self contact between parts of the body.

\subsection{\methodname Wnet}\label{ssec:WBR}

Here we describe \methodname Wnet (BPWnet), a deep network with a white-box model of depth and pressure image generation, shown in Fig.~\ref{fig:networks}. BPWnet uses a traditional black-box CNN for encoding depth images, but uses a white-box model with no learned parameters for reconstructing depth and pressure images from an embedded SMPL human model. It contains two modules: `Mod1', which produces an initial estimate, and `Mod2', which refines the estimate through residual error in a similar way to previous works~\cite{oberweger2015training, carreira2016human, clever2020bodies}. 

First, BPWnet maps a depth image $\mathcal{D}$ to SMPL parameters $\boldsymbol{\widehat{\Psi}}$ using a CNN. From there, it uses the differentiable SMPL embedding from Kanazawa \etal~\cite{kanazawa2018end} to produce a SMPL mesh estimate, $\widehat{\mathcal{M}}_H$. A loss is applied using height and weight estimates from BetaNet. The first module, Mod1, contains a white-box model of depth image generation, which differentiably reconstructs depth maps $\widehat{\mathcal{D}}^+$ from a SMPL mesh. The spatial residual between these maps and the input depth images are used to learn a correction and refine initial human pose estimates in the second module, Mod2. In contrast to Mod1, Mod2 contains a white-box model of pressure image generation, which differentiably reconstructs pressure maps $\widehat{\mathcal{P}}^+$ from an improved SMPL mesh estimate. Finally, the CAL component in Fig.~\ref{fig:networks} adjusts pressure maps to achieve a similar calibration to real pressure images.

\textbf{Depth image encoding.} Both modules of BPWnet encode depth imagery with ResNet34 convolutional neural networks (CNNs). Each CNN outputs estimated SMPL parameters $\widehat{\boldsymbol{\Psi}} = \big[ \boldsymbol{\hat{\beta}} \hspace{3mm} \boldsymbol{\widehat{\Theta}} \hspace{3mm} \boldsymbol{\hat{s}} \hspace{3mm} \boldsymbol{\hat{x}} \hspace{3mm} \boldsymbol{\hat{y}} \hspace{3mm} \hat{b} \big]^{\top} \in \mathbb{R}^{89}$, which contains body shape, joint angles, and root translation and rotation. It also contains an estimated distance between the camera and the bed, $\hat{b}$, which is described later in this section. The SMPL parameters are used to compute a SMPL human mesh model with no learnable weights. The SMPL block takes as additional input a set of gender flags $\boldsymbol{g} \in \mathbb{R}^2$ (recall Section~\ref{ssec:betanet}) and outputs a human mesh $\widehat{\mathcal{M}}_H$. We define a loss on the SMPL model, $\mathcal{L}_{\textrm{SMPL}}$, which minimizes error from the SMPL parameters and 3D SMPL joint positions. We also define a loss on the SMPL vertex positions, $\mathcal{L}_{\mathrm{v2v}}$, which can be used in conjunction with $\mathcal{L}_{\textrm{SMPL}}$ to provide more supervision at a marginally higher computational cost.  
Appendix~\ref{sec:WBR_details} contains details on $\mathcal{L}_{\textrm{SMPL}}$ and $\mathcal{L}_{\mathrm{v2v}}$. 
BPWnet also provides supervision on human mass and height using the BetaNet described in Section~\ref{ssec:betanet}.

\begin{figure}
\begin{center}
\centering
\includegraphics[width=8.5cm]{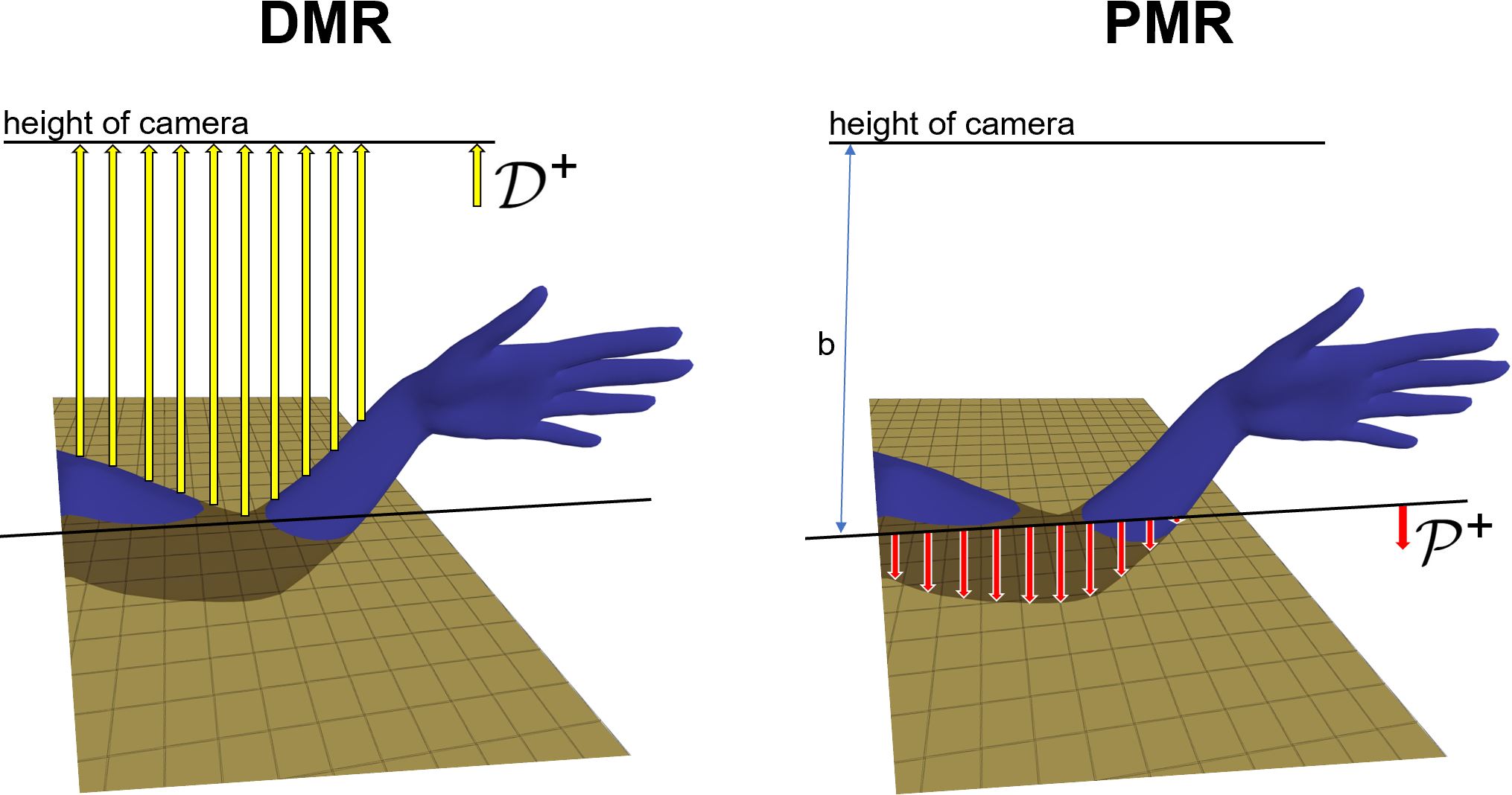}
\vspace{-4mm}
\end{center}
\caption{Differentiable white-box depth and pressure map reconstruction. DMR computes a linear depth map $\mathcal{D}^+$ between the height of the camera and the top surface of the human mesh (left). PMR computes a linear pressure map $\mathcal{P}^+$ between the undeformed height of the surface of the bed and the human mesh (right). Variable $b$ is the distance between the camera and the bed.}

\label{fig:dmr_pmr_calc}
\end{figure}

\textbf{White-box decoding.} In Mod1, the depth map reconstruction (DMR) module computes a depth map $\mathcal{\widehat{D}}^+ \in \mathbb{R}^{64 \times 27}$ from the body mesh $\widehat{\mathcal{M}}_{H,1}$. This is computed by calculating the distance between the camera plane and the inferred human mesh (Fig.~\ref{fig:dmr_pmr_calc} - left). 
The reconstructed depth map is used for residual error refinement in Mod2: The difference between the real pose in the input depth image $\mathcal{D}$ and the pose in the reconstructed depth map $\mathcal{\widehat{D}}^+$ contains information that can be used to improve the initial pose and body shape estimate. This white-box model of depth image generation is similar to the pressure image generation introduced in our previous work~\cite{clever2020bodies}, and is differentiable so the output can be used during training. Unlike the input depth image $\mathcal{D}$, the DMR reconstruction $\mathcal{\widehat{D}}^+$ only contains human mesh information and is occlusion-free; specifically, it does not contain blanket or mattress information because its sole purpose is to improve the initial pose estimate.

Mod2 uses pressure map reconstruction (PMR) from~\cite{clever2020bodies} to compute a pressure map $\mathcal{\widehat{P}}^+ \in \mathbb{R}^{64 \times 27}$ from the refined pose estimate $\widehat{\mathcal{M}}_{H,2}$. 
This is the distance the body mesh sinks into the underlying mattress (Fig.~\ref{fig:dmr_pmr_calc} - right). In contrast to the previous work that reconstructs a pressure map from a \textit{pressure} image, in BPWnet, PMR reconstructs a pressure map from a \textit{depth} image.  
The position of the mattress must be known with high accuracy relative to the depth camera,  
because the pressure is sensitive to small changes in the vertical distance between the camera and bed. Small camera movements or the weight of a large person on the bed can change the perceived distance enough to alter the reconstructed pressure map. Thus, it requires an additional variable to represent changes in the vertical distance between the camera and the surface of the bed, which can correct for changes in the camera's position. We define this parameter as $b$, and learn it from depth images at varying distance from the bed. We also define a loss on the PMR component, $\mathcal{L}_{\mathcal{P}^{+}}$, which can be used to train the depth image encoder. See details in Appendix~\ref{sec:WBR_details}.

\textbf{Calibrating the inferred pressure map.}
While $\mathcal{\widehat{P}}^+$ is spatially similar to a pressure image, it has some qualitative differences: it contains more dilated features, it has less noise, and the magnitude of each pixel is a distance rather than a pressure. Thus, our approach uses a small convolutional network, CAL, to calibrate $\mathcal{\widehat{P}}^+$, converting it to $\mathcal{\widehat{P}} \in \mathbb{R}^{64 \times 27}$. CAL takes as input a stack of 3 images including $\mathcal{\widehat{P}}^+$ and constant CoordConv maps $\mathcal{R} \in \mathbb{R}^{2\times 64\times 27}$,  
which allow the network to model non-translation invariant aspects and can improve trainability and generalization~\cite{liu2018intriguing}. CAL contains 4 layers of convolution, 
with $< 0.4\%$ as many parameters as the encoder. We define a loss based on the output pressure image, $\mathcal{L}_{\mathcal{P}}$, which can be used to train CAL and is detailed in Appendix~\ref{sec:WBR_details}.

\subsubsection{Training Strategy} 
We train the encoders for Mod1 and Mod2 separately. The loss for training the learnable weights in Mod1 is computed as:

\begin{equation}
\mathbb{L}_{\textrm{BPW}_1} = \mathcal{L}_{\textrm{BetaNet}_1}+\mathcal{L}_{\textrm{SMPL}_1}
\label{eqn:WB1}
\end{equation}

where BetaNet is separately pretrained (Section~\ref{ssec:betanet}) and contains frozen network weights. Then, the entire dataset is passed forward through the network to compute a set of Mod1 estimates with each sample containing $\{ \boldsymbol{\widehat{\Psi}}_{1}, \widehat{\mathcal{D}}^+, \widehat{\mathcal{C}}_{\mathrm{d}^+} \}$, where $\widehat{\mathcal{C}}_{\mathrm{d}^+}$ are the binary contact maps of $\widehat{\mathcal{D}}^+$. 
In this forward pass, noise is added to the SMPL parameters to increase the variation in the types of error that Mod2 corrects. With a dataset consisting of inputs $\{ \boldsymbol{\mathcal{D}}, \boldsymbol{\widehat{\mathcal{D}}}^+, \boldsymbol{\widehat{\mathcal{C}}}_{\mathrm{d}^+}\}$, we train the encoder in Mod2 to learn a residual correction $(\boldsymbol{\widehat{\Psi}}_{2} - \boldsymbol{\widehat{\Psi}}_{1})$ with the following loss function:

\begin{equation}
\mathbb{L}_{\textrm{BPW}_2} = \mathcal{L}_{\textrm{BetaNet}_2}+\mathcal{L}_{\textrm{SMPL}_2}+\mathcal{L}_{\mathrm{v2v}_2}+\mathcal{L}_{\mathcal{P}^{+}_2}
\label{eqn:WB2}
\end{equation}

After training the depth image encoders, we train the CAL network. CAL learns to refine the features of $\widehat{\mathcal{P}}^+$ and calibrate pressure values at individual taxels rather than to spatially adjust pressure for a change in limb or body movement. Ground truth human meshes from the dataset are used to compute ground truth reconstructed pressure maps $\mathcal{P}^+$, which are fed into CAL during training. CAL outputs the estimate $\widehat{\mathcal{P}}$, which is compared to ground truth $\mathcal{P}$ during training, using loss $\mathcal{L}_{\mathcal{P}}$.




\section{Evaluation}\label{sec:evaluation}

We evaluate our method using the SLP multimodal dataset~\cite{liu2020simultaneously}, which is a human pose dataset consisting of $4,590$ unique resting poses in bed across $102$ human participants. Each pose is captured with three different situations of varying visual occlusion: (1) thin sheet covering the person, (2) thicker blanket covering, and (3) no covering. The dataset contains RGB, depth, point cloud, thermal, and pressure imagery, as well as 2D human pose keypoints; the bottom row of Fig. 1 (b) and Fig.~\ref{fig:slp_fitting} provide a couple examples. In Section~\ref{ssec:bodypressure_sd}, we describe the data partitions generated using the method in Section~\ref{sec:methods1}. Finally, in Section~\ref{ssec:P_eval}, we describe the evaluation of the deep network from Section~\ref{sec:methods2} on the SLP dataset.

\subsection{\methodname SD Synthetic Dataset Partitions}\label{ssec:bodypressure_sd}

The synthetic data generation method from Section~\ref{sec:methods1} is used to generate \methodname SD: a large collection of samples, each of which includes a resting pose, a unique body shape, a gender, a depth image, a pressure image, and four associated meshes from the scene for the person, mattress, pressure sensing mat, and blanket. The pressure images for both this data and the real SLP data are normalized by body mass. The depth and pressure images are spatially co-registered with the calibrated images in the real SLP dataset. Details on these procedures are provided in Appendices~\ref{sec:massnorm} and~\ref{sec:appendix_coreg}.

\begin{table}[t!]
\caption{Human Pose and Body Shape Dataset Partitions}
\label{tab:partition_descrip}
\begin{center}
\vspace{-6mm}
\footnotesize

\renewcommand{\arraystretch}{1.1}
\vspace{1mm}

\scalebox{0.9}{\begin{tabular} {c|c|c|c}
\hline

 & synthetic & real & real \\
Description & train ct. & train ct. & test ct.\\
\hline
\hline
Supine - Unique Images w Blankets & 34619 & 2370 & 660 \\
L. Lateral - Unique Images w Blankets & 32050 & 2370 & 660 \\
R. Lateral - Unique Images w Blankets & 30826 & 2370 & 660 \\
Supine - Unique Images w/o Blankets & 34619 & 1185 & 330 \\
L. Lateral - Unique Images w/o Blankets & 32050 & 1185 & 330 \\
R. Lateral - Unique Images w/o Blankets & 30826 & 1185 & 330 \\
\hline
Total Unique Images & 194990 & 10665 & 2970 \\
\hline
\hline
Supine - Unique Poses & 34619 & 1185 & 330 \\
Right Lateral - Unique Poses & 32050 & 1185 & 330 \\
Left Lateral - Unique Poses & 30826 & 1185 & 330 \\
\hline
Total Unique Poses & 97495 & 3555 & 990 \\
\hline
\hline
Total Unique Body Shapes & 97495 & 79 & 22 \\
\hline
\hline
Num Samples for Training and Testing & 97495 & 10665 & 2970 \\
\hline

\end{tabular}}
\end{center}
\end{table}

\textbf{Human body pose partitions.} To create the synthetic training dataset, our process samples initial poses and body shapes close to the real poses as depicted in Fig.~\ref{fig:pipeline}-I. For each unique real pose among the $80$ training subjects in the SLP dataset, it attempts to generate $30$ initial synthetic poses split evenly between female and male SMPL bodies. Each of these poses have a unique body shape.  
Given $80$ participants in the training dataset with $45$ unique poses per participant, this would ideally generate $30 \times 80 \times 45 = 108,000$ initial poses and body shapes. 
However, in some cases it is challenging to find a collision-free pose for a particular body shape that is close to a particular real pose, so some samples are aborted when a limit is reached. This reduces the initial pose/body shape count to $103,966$.


Next, the process runs this set of bodies with initial poses through Physics Simulations \#1 and \#2, of which some more are rejected due to the simulation becoming unstable. The process is designed to automatically detect when simulation instability is imminent, in which case it aborts the simulation and rejects the pose. This can happen due to situations such as a limb poking a hole in the pressure mat, which are described in our previous work~\cite{clever2020bodies}. None of the blanket covering simulations resulted in instability. This resulted in a total of $97,495$ unique data samples, which are used to train the network. The data partitions are broken down in Table~\ref{tab:partition_descrip}. 

\textbf{Blanket configuration partitions.} 
The SLP dataset contains both thick and thin covers, which are placed to cover most of the body. The simulator is only equipped to generate blankets with a single fixed thickness, which we assume is close enough to represent both real coverings. The real blankets often contain many wrinkle features that may be caused by pulling or adjusting the blanket so that it covers the body appropriately. We attempt to mimic these situations. The process incorporates randomization in the initial position of the blanket over the surface of the bed $\boldsymbol{s}_B$ (recall the top half of Fig.~\ref{fig:pipeline}-V). 
Each resting body is associated with a single initial blanket position and the covering variation that results from it. The initial blanket configurations are split into two partitions. In the first, the blanket is centered over the person with the upper edge coinciding with the person's neckline, such that no pulling is required to uncover the head or cover the body (recall Fig.~\ref{fig:pipeline}-VI.).  In the second partition, the initial blanket position is randomly sampled across the person in bed. These two blanket configuration scenarios are split 50/50 among the synthetic data pose partitions described in Table~\ref{tab:partition_descrip}. 
In all cases, the initial blanket rotation, $\boldsymbol{\phi}_B$, is set to a constant value of $\boldsymbol{0}$. Appendix~\ref{sec:app_blanket_partition} provides details on the specific sampling bounds. 

\begin{figure*}
\begin{center}
\centering
\includegraphics[width=17.5cm]{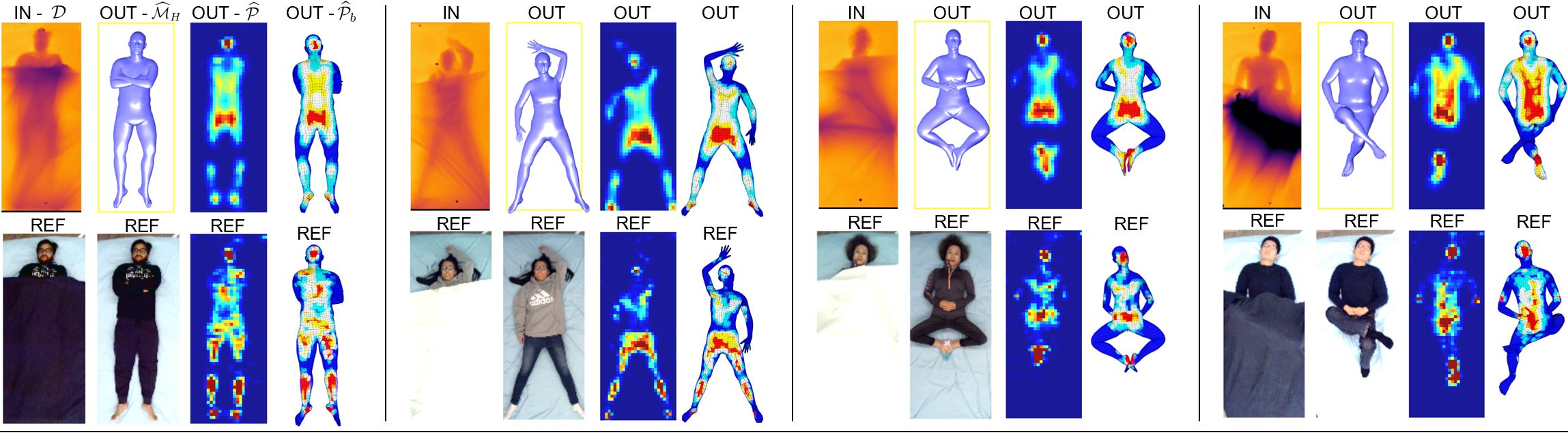}
\includegraphics[width=17.5cm]{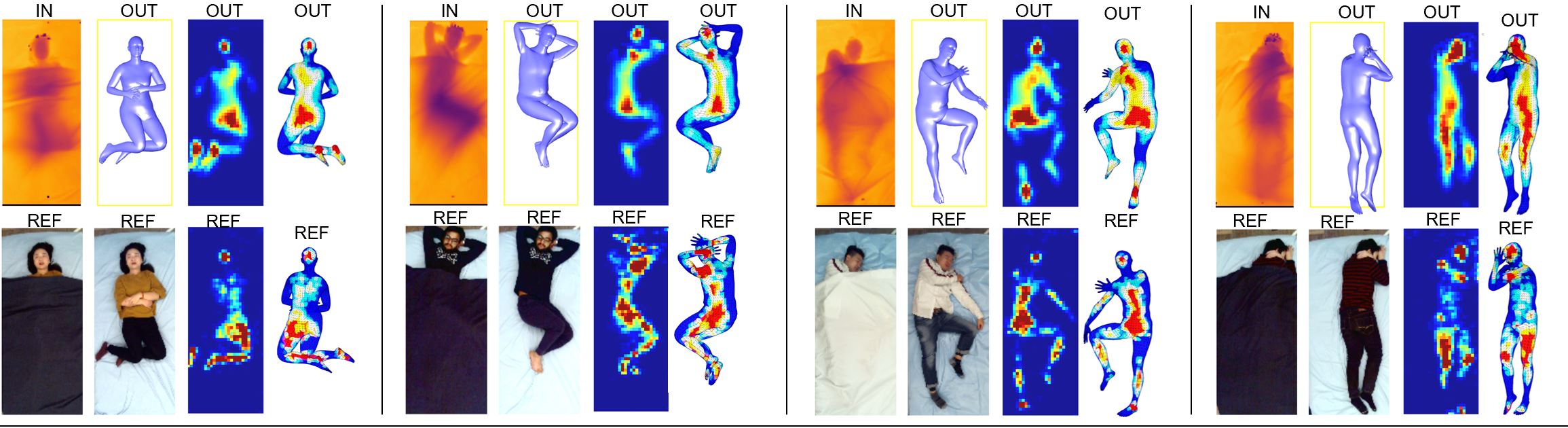}
\includegraphics[width=17.5cm]{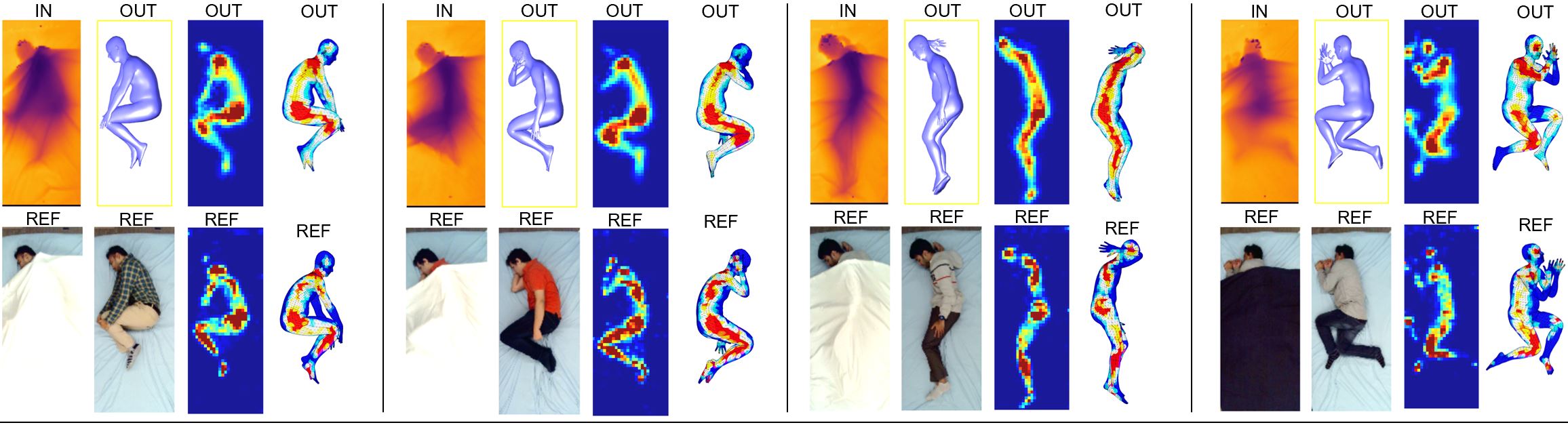}
\includegraphics[width=17.5cm]{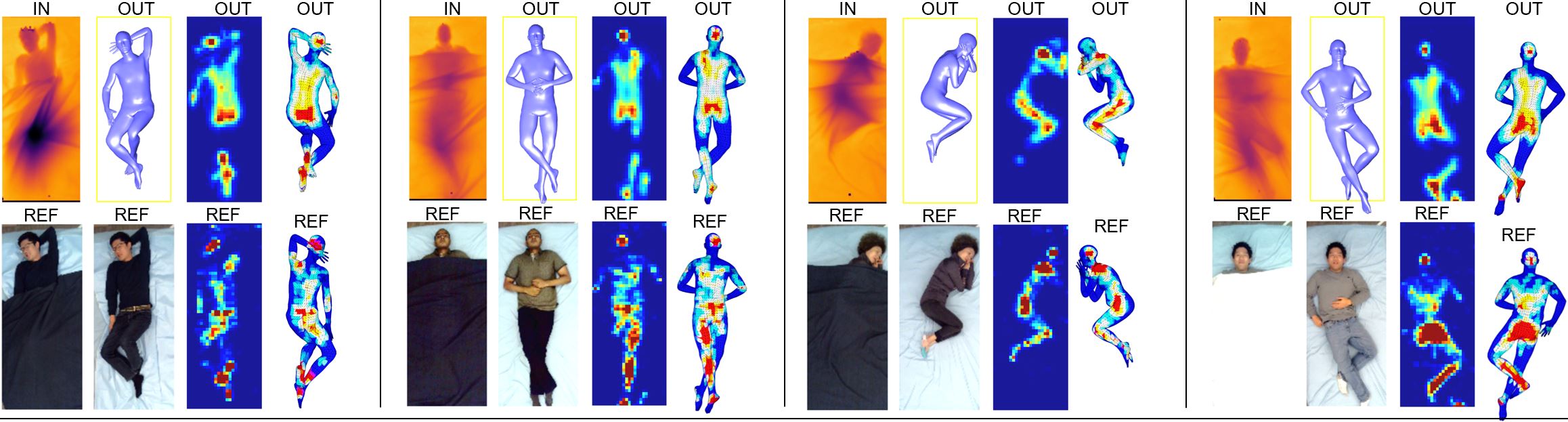}
\includegraphics[width=17.5cm]{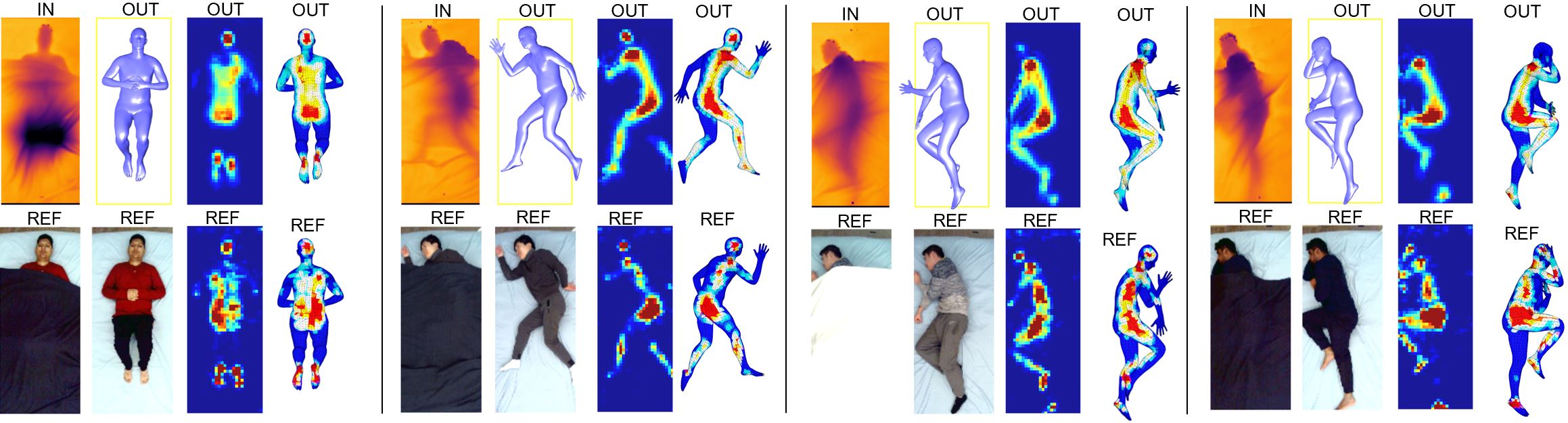}
\end{center}
\vspace{-0.6cm}
\caption{Results: Inferring pose, contact pressure, and localized pressure distribution from depth using BPWnet. Showing real data with people occluded by blankets in the depth images. All cases are from the last 22 test subjects in the SLP dataset; white coverings indicate `cover 1' and black indicate `cover 2.' The far right renderings in each group are a mirror flip because they show the pressure distribution \textit{underneath} the body; the top shows an inferred pose while the bottom shows a pose from the SLP-3Dfits annotations.}
\label{fig:results_fig_main}
\end{figure*}

\subsection{Network Evaluation}\label{ssec:P_eval}

For human pose inference, we compare our method directly to the Pyramid fusion scheme by Yin \etal~\cite{yin2020multimodal}, which uses 4 modalities (RGB, depth, thermal, and pressure imagery) to infer human pose in the form of a SMPL mesh. Our method uses only depth imagery, which would have significant advantages for real-world deployment. Pose accuracy is evaluated with 3D mean-per-joint position error (MPJPE), using the same 22-subject test set proposed in Yin \etal. 
For each pose sample, the inferred positions of 24 joints on the SMPL model are compared to ground truth using 3D Euclidean error.

For both contact pressure inference and the inferrence of pressure on the human body, we did not find any existing work for comparison. We designed a second deep architecture to compare BPWnet with, which uses a more traditional black-box method of image reconstruction. We refer to this alternative as \methodname Bnet (BPBnet). BPBnet replaces the white-box DMR and PMR components in BPWnet with black-box decoders. These learned image decoders are designed symmetrically to the encoders, expanding the SMPL parameters back into images. 
Appendix~\ref{sec:resusmpl} contains details about BPBnet. We train both network architectures with the same hyperparameters, and compare the inferred pressure image to ground truth using mean-squared error (MSE), which is computed on a per-taxel basis. We compare the ability to localize regions of high pressure density using vertex-to-vertex pressure (v2vP) mapped to the SMPL model, where MSE is computed on a per-vertex basis. Because vertices are not evenly distributed over the surface of the body and pressure is in units of $force / area$, the pressure on each vertex is normalized by the average area of the adjacent triangles. We also compare human pose estimation error between BPWnet and BPBnet.

\textbf{Dataset splits.} We trained our networks on datasets consisting of real, synthetic, and combined synthetic and real data. For the synthetic training data, we selected depth images with blanket occlusions on $2/3$ of the poses, and depth images without blankets for $1/3$ of the poses. This matches the real data, of which $2/3$ of images are occluded by blankets $1/3$ are not. We performed validation and testing on real data. We created both training/validation and training/testing splits based on the $102$ subjects in the SLP dataset. For training/validation, we trained on data from the first 70 subjects (i.e. subjects 1 - 70, with $9,315$ real and $85,114$ synthetic samples), and validated on the next 10 subjects (i.e. subjects 71 - 80, with $1,350$ real samples). We used this split for tuning network hyper parameters. For training/testing, we used the same split as Yin \etal ~\cite{yin2020multimodal}, and trained on data from the first 80 subjects (i.e. subjects 1 - 80, with $10,665$ real and $97,495$ synthetic samples), and tested on the last 22 subjects (i.e. subjects 81 - 102, with $2,970$ real samples). We did not use subject 7 data due to errors in calibration. 

\begin{table*}[t!]
\caption{\label{tab:results_table_pose}Human pose estimation (Pose), contact pressure (P. Img.), and pressure distribution (v2vP) error results when evaluating on the 22 subject test set.}
\vspace{-5mm}
\begin{center}

\renewcommand{\arraystretch}{1.1}

\scalebox{0.95}{\begin{tabular} {c|c|c|c|c||c|c|c|c||c|c|c|c||c|c|c|c}
\hline
 \hspace{-2mm} Network Description \hspace{-2mm}&   
 \parbox[t]{2mm}{{\rotatebox[origin=c]{90}{Training}}}
 \parbox[t]{2mm}{{\rotatebox[origin=c]{90}{data ct.}}}&
 \parbox[t]{1mm}{{\rotatebox[origin=c]{90}{Real}}}&  
 \parbox[t]{1mm}{{\rotatebox[origin=c]{90}{Synthetic}}}&  
 \parbox[t]{3mm}{{\rotatebox[origin=c]{90}{Modalities}}}&
 \parbox[t]{3mm}{{\rotatebox[origin=c]{90}{Uncovered - Pose }}}
 \parbox[t]{2mm}{{\rotatebox[origin=c]{90}{MPJPE (mm)}}}&
 \parbox[t]{3mm}{{\rotatebox[origin=c]{90}{Cover 1 - Pose }}}
 \parbox[t]{2mm}{{\rotatebox[origin=c]{90}{MPJPE (mm) }}}&
 \parbox[t]{3mm}{{\rotatebox[origin=c]{90}{Cover 2 - Pose }}}
 \parbox[t]{2mm}{{\rotatebox[origin=c]{90}{MPJPE (mm) }}}& 
 \parbox[t]{3mm}{{\rotatebox[origin=c]{90}{Overall - Pose }}}
 \parbox[t]{2mm}{{\rotatebox[origin=c]{90}{MPJPE (mm) }}}&
 \parbox[t]{3mm}{{\rotatebox[origin=c]{90}{Uncovered - P. Img.}}}
 \parbox[t]{2mm}{{\rotatebox[origin=c]{90}{MSE (kPa$^2$)}}}&
 \parbox[t]{3mm}{{\rotatebox[origin=c]{90}{Cover 1 - P. Img.}}}
 \parbox[t]{2mm}{{\rotatebox[origin=c]{90}{MSE (kPa$^2$) }}}&
 \parbox[t]{3mm}{{\rotatebox[origin=c]{90}{Cover 2 - P. Img.}}}
 \parbox[t]{2mm}{{\rotatebox[origin=c]{90}{MSE (kPa$^2$) }}}& 
 \parbox[t]{3mm}{{\rotatebox[origin=c]{90}{Overall - P. Img.}}}
 \parbox[t]{2mm}{{\rotatebox[origin=c]{90}{MSE (kPa$^2$) }}}&
 \parbox[t]{3mm}{{\rotatebox[origin=c]{90}{Uncovered - v2vP}}}
 \parbox[t]{2mm}{{\rotatebox[origin=c]{90}{MSE (kPa$^2$)}}}&
 \parbox[t]{3mm}{{\rotatebox[origin=c]{90}{Cover 1 - v2vP}}}
 \parbox[t]{2mm}{{\rotatebox[origin=c]{90}{MSE (kPa$^2$) }}}&
 \parbox[t]{3mm}{{\rotatebox[origin=c]{90}{Cover 2 - v2vP}}}
 \parbox[t]{2mm}{{\rotatebox[origin=c]{90}{MSE (kPa$^2$) }}}& 
 \parbox[t]{3mm}{{\rotatebox[origin=c]{90}{Overall - v2vP}}}
 \parbox[t]{2mm}{{\rotatebox[origin=c]{90}{MSE (kPa$^2$) }}}\\

\hline
\hline
BPWnet & 97K & &\hspace{-2mm} $\times$ & D & \hspace{-2mm} 95.79\hspace{-1mm} & \hspace{-2mm} 112.08\hspace{-1mm} & \hspace{-2mm} 109.83\hspace{-1mm} & \hspace{-2mm} 105.90\hspace{-1mm} & \hspace{-2mm} 1.665\hspace{-1mm} &		 \hspace{-2mm} 1.764\hspace{-1mm} & \hspace{-2mm} 1.738\hspace{-1mm} & \hspace{-2mm} 1.772\hspace{-1mm} & \hspace{-2mm} 3.861\hspace{-1mm} & \hspace{-2mm} 3.839\hspace{-1mm} & \hspace{-2mm} 3.812\hspace{-1mm} & \hspace{-2mm} 3.837\hspace{-1mm} \\
BPWnet & 11K &\hspace{-2mm} $\times$ &\hspace{-2mm} & D & \hspace{-2mm} 103.22\hspace{-1mm} & \hspace{-2mm} 104.56\hspace{-1mm} & \hspace{-2mm} 104.83\hspace{-1mm} & \hspace{-2mm} 104.20\hspace{-1mm} & \hspace{-2mm} 1.470\hspace{-1mm} & \hspace{-2mm} 1.455\hspace{-1mm} & \hspace{-2mm}	 1.444\hspace{-1mm} & \hspace{-2mm} 1.456\hspace{-1mm} & \hspace{-2mm}		 3.192\hspace{-1mm} & \hspace{-2mm} 3.148\hspace{-1mm} & \hspace{-2mm} 3.112\hspace{-1mm} & \hspace{-2mm} 3.151\hspace{-1mm} \\
Pyramid Fusion~\cite{yin2020multimodal} \hspace{-1mm} & 11K &\hspace{-2mm} $\times$ &\hspace{-2mm} &\hspace{-2mm} RGB-D-T-P\hspace{-1mm} & \hspace{-2mm} 78.80\hspace{-1mm} & \hspace{-2mm} 79.92\hspace{-1mm} & \hspace{-2mm} 80.21\hspace{-1mm} & \hspace{-2mm} 79.64\hspace{-1mm} & - & - & - & - & - & - & - & -  \\
BPBnet, No SMPL & 108K &\hspace{-2mm} $\times$ &\hspace{-2mm} $\times$ & D & \hspace{-2mm} - & - & - & - & \hspace{-2mm} 0.825\hspace{-1mm} & \hspace{-2mm} 0.959\hspace{-1mm} & \hspace{-2mm} 0.932\hspace{-1mm} & \hspace{-2mm} 0.905\hspace{-1mm} & - & - & - & - \\
BPBnet & 108K &\hspace{-2mm} $\times$ &\hspace{-2mm} $\times$ & D & \hspace{-2mm} 70.16\hspace{-1mm} & \hspace{-2mm} 76.99\hspace{-1mm} & \hspace{-2mm} 76.49\hspace{-1mm} & \hspace{-2mm} 74.54\hspace{-1mm} & \hspace{-2mm} \textbf{0.772}\hspace{-1mm} & \hspace{-2mm} \textbf{0.884}\hspace{-1mm} & \hspace{-2mm} \textbf{0.858}\hspace{-1mm} & \hspace{-2mm} \textbf{0.838}\hspace{-1mm} & \hspace{-2mm} 2.872\hspace{-1mm} & \hspace{-2mm} 2.877\hspace{-1mm} & \hspace{-2mm} 2.868\hspace{-1mm} & \hspace{-2mm} 2.872\hspace{-1mm} \\
BPWnet & 108K$\dagger$ &\hspace{-2mm} $\times$ &\hspace{-2mm} $\times$ & D & \hspace{-2mm} \textbf{63.64}\hspace{-1mm} & \hspace{-2mm} \textbf{72.40}\hspace{-1mm} & \hspace{-2mm} \textbf{72.04}\hspace{-1mm} & \hspace{-2mm} \textbf{69.36}\hspace{-1mm} & \hspace{-2mm} 1.155\hspace{-1mm} & \hspace{-2mm} 1.209\hspace{-1mm} & \hspace{-2mm} 1.190\hspace{-1mm} & \hspace{-2mm} 1.184\hspace{-1mm} & \hspace{-2mm}			 \textbf{2.854}\hspace{-1mm} & \hspace{-2mm} \textbf{2.860}\hspace{-1mm} & \hspace{-2mm} \textbf{2.832}\hspace{-1mm} & \hspace{-2mm} \textbf{2.849}\hspace{-1mm} \\
\hline
\cline{1-17}

\end{tabular}}
\end{center}
{ $\dagger$ \textit{indicates the encoder was trained with 108K mixed, while the CAL component was trained using only 11K real.}}
\end{table*}

\textbf{Depth image noise model.} Our camera noise model includes white noise, dropout, and synthetic occlusion on sections of the input image using the code provided by Liu \etal ~\cite{liu2020simultaneously}. Because the inferred contact pressure is highly sensitive to the distance between the camera and the bed, 
a single distance is uniformly sampled between -5 and 5 cm for each image, which is added to all depth pixels to make the inference robust to vertical movements of the camera or bed. No rotational noise or translational noise in the plane normal to gravity are added, since they appeared to be unnecessary for the SLP dataset. Noise is also added to account for the physics of the bed springs underneath the soft mattress. The simulated mattress is set on a rigid plane, which differs from the flexible springs in Invacare Homecare bed used to collected real data in~\cite{clever2020bodies} as well the bed used to collect the SLP dataset~\cite{liu2020simultaneously}. In practice, we observe a substantial drop in the middle of the bed when a person rests on it. This is modeled with a 2D parabolic map added to depth images during training, which is equal to zero at the edges of the bed and increases to a max in the center. A parameter that alters this max value is uniformly sampled between 0 and 10 cm.


\textbf{Network hyper-parameters.} For all networks, we shuffled the training data, used a batch size of 128, used the ADAM optimizer~\cite{kingma2014adam} for gradient computation, and used a learning rate of $0.0001$ and weight decay of $0.0005$. For BetaNet, we trained for $500$ epochs on real and synthetic data. The BetaNet used in the optimization of Section~\ref{sec:slp_optim} was only trained with synthetic data because the body shape parameters were not available prior to annotation. 
For the CAL network in BPWnet, we trained for $500$ epochs on real and synthetic data. For both BPBnet and BPWnet, we trained on $100$ epochs on the first module. Then, we pre-computed their estimates and used it for training the second module, which we trained for $40$ epochs. We trained for $40$ epochs because the network began to overfit the training data at this point regardless of the training dataset used. Our machine has a AMD Ryzen Threadripper 1950X 16-Core processor with 64 GB of CPU RAM and a NVIDIA RTX-3090 GPU. Training each network module took $\sim$12 hrs. Overall, BPWnet had better computational performance than BPBnet, which is detailed in Appendix~\ref{sec:computation}.


\section{Results and Discussion}\label{sec:results}

In this section, we present and discuss the results of the evaluation. 

\textbf{Our method for estimating pose outperforms the state-of-the-art.} 
Using only a depth image in the input, BPWnet is able to infer pose with 12\% lower error than the state-of-the-art method from~\cite{yin2020multimodal}, which uses a combination of RGB, depth, thermal, and pressure imagery to infer pose.  BPWnet also outperformed the alternative BPBnet. Table~\ref{tab:results_table_pose} presents these results, with comparisons between the type of covering on the person. Fig.~\ref{fig:results_fig_main} presents a visual overview of BPWnet.

\textbf{Mixing synthetic and real data boosts performance.} When training the network only on real depth images from the SLP dataset or only on synthetic depth images from \methodname SD, performance lags. However, when the real and synthetic datasets are naively mixed into a single bag of training data, pose error drops by more than 30\%.

\textbf{Our method can infer contact pressure from depth.}
Our method can infer a pressure image ($\mathcal{P}$) from overhead depth imagery, which is also depicted visually in Fig.~\ref{fig:results_fig_main}. Like pose inference, using a mixed bag of synthetic and real data boosts performance, as shown in Table~\ref{tab:results_table_pose}. The error for BPBnet is significantly lower than BPWnet. When the SMPL model is ablated from BPBnet, it is still able to infer pressure but performs worse. This indicates that joint learning of the SMPL parameters helps BPBnet learn contact pressure from depth. We did not ablate the SMPL model from BPWnet because it requires SMPL to infer a pressure image.

\begin{figure}
\begin{center}
\centering
\includegraphics[width=8.8cm]{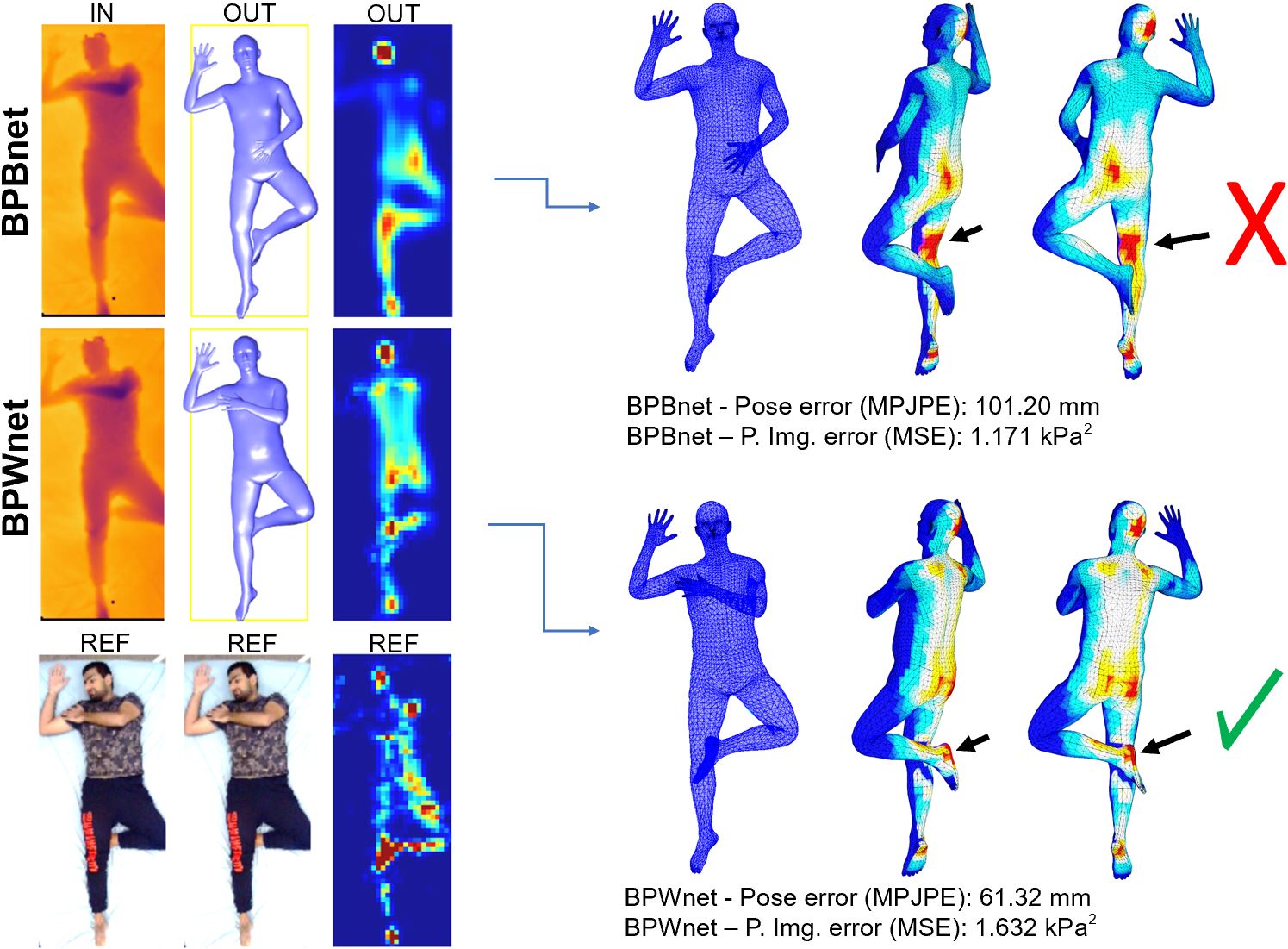}
\end{center}
\vspace{-0.5cm}
\caption{Behavior comparison of BPBnet and BPWnet. BPBnet has lower contact pressure error, but its projection onto the inferred mesh contains an artefact. BPWnet plausibly infers a high pressure on the foot, while BPBnet incorrectly assigns a high pressure to the underside of the upper leg.}
\label{fig:WBR_resusmpl_comparison}
\vspace{-0.2cm}
\end{figure}

\textbf{BPWnet can infer pressure on the body.} While black-box image reconstruction is far more common in computer vision and our black-box model (BPBnet) has a lower pressure image inference error, our white-box model (BPWnet) has a lower pressure distribution ($\mathcal{P}_b$) error as measured by vertex-to-vertex pressure (v2vP) in Table~\ref{tab:results_table_pose}. Generally, BPWnet can more reliably localize pressure. This is because the inferred pressure image in BPWnet can be reduced to a function of only the SMPL parameters (i.e. $\widehat{\mathcal{P}} = f(\boldsymbol{\widehat{\Psi}})$ ) where the spatial mapping between the human mesh $\widehat{\mathcal{M}}_H$ and the pressure map $\widehat{\mathcal{P}}^+$ is a known geometric function with no learnable parameters, so the reconstructed pressure image reliably projects onto the surface of the inferred human mesh. In contrast, there is little in the black-box model to ensure the inferred pressure map spatially co-registers with the inferred mesh. We present a visual example of this phenomena in Fig.~\ref{fig:WBR_resusmpl_comparison}. In it, the person lays supine on the bed with the left foot tucked under the right upper leg. This produces a peak pressure on the left foot, which carries extra weight from the right leg. BPWnet appropriately projects the peak pressure onto the left foot, while the BPBnet projection contains a discrepancy in the peak pressure location.

\textbf{Pose vs. contact pressure accuracy tradeoff.} A number of factors may account for the tradeoff in pose vs. contact pressure accuracy between BPWnet and BPBnet. For pose estimation, the reconstructed depth map estimate $\widehat{\mathcal{D}}^+$ in BPWnet is a constrained function of SMPL parameters and thus provides more consistent spatial residual feedback to learn the pose correction in Mod2. If the input depth image is highly occluded or contains noise, BPWnet may produce a poor initial pose estimate but $\widehat{\mathcal{D}}^+$ will contain no less pose information. In contrast, the black-box reconstructed depth map in BPBnet has no lower bound on the amount of information it contains, so in some cases it may not contain any useful information for the correction. 

For the pressure image inference, Table~\ref{tab:results_table_pose} indicates that the SMPL model is not necessary for inferring the pressure image with BPBnet. Thus, a pixel-to-pixel black-box network is sufficient for inferring a pressure image from an occluded depth image. BPWnet likely decreases the performance of this inference because it has fewer free parameters and a tighter grasp on the pressure image formation process: if the pose changes, the pressure necessarily changes. Some poses may be more challenging to learn than some instances of contact pressure, so an inaccurate pose would adversely affect the pressure image inference.

\textbf{Ablating BPWnet components reduces the performance of body pressure inference.} 
We conducted an ablation study to test the importance of components in BPWnet, shown in Table~\ref{tab:results_table_ablation}. We ablated BetaNet, which improved pose accuracy, marginally reduced pressure accuracy, and substantially reduced accuracy of body height and weight. Coincidentally, the BetaNet model of height and mass also seems to cause a tradeoff in pose vs. pressure accuracy. However, the reduction in overall body shape accuracy as measured by height and weight casts some doubt on the merit of omitting BetaNet. We ablated the CAL feature calibration component, which affects only the contact pressure inference. Without CAL, the pressure inference performs poorly because the overall scale of the PMR output, $\mathcal{P}^+$ is different than $\mathcal{P}$. This indicates that CAL is able to scale the pressure from an arbitrary range to an appropriate range. We also ablated the residual learning by removing the DMR component in Mod1 and used PMR instead to compute pressure maps. For this, we trained only the initial CNN for 100 epochs but used the $\mathbb{L}_{\textrm{BPW}_2}$ loss. This marginally changed pose, pressure, and body height error but substantially reduced the accuracy of body weight.

\begin{table}[t!]
\caption{\label{tab:results_table_ablation} Ablation study - evaluated on the 22 subject test set. Pose and pressure error shown with same metrics as previous tables. Body mass and height are evaluated with mean absolute error.}
\vspace{-6mm}
\begin{center}
\renewcommand{\arraystretch}{1.1}
\scalebox{0.89}{\begin{tabular} {c|c|c|c|c|c|c|c|c}
\hline
 Network &    
 \parbox[t]{2mm}{{\rotatebox[origin=c]{90}{Residual}}}  
 \parbox[t]{1mm}{{\rotatebox[origin=c]{90}{w/ DMR}}}&
 \parbox[t]{2mm}{{\rotatebox[origin=c]{90}{BetaNet}}}&
 \parbox[t]{2mm}{{\rotatebox[origin=c]{90}{CAL feature}}}
 \parbox[t]{1mm}{{\rotatebox[origin=c]{90}{calibration}}}&
 \parbox[t]{2mm}{{\rotatebox[origin=c]{90}{Overall - Pose}}}
 \parbox[t]{1mm}{{\rotatebox[origin=c]{90}{MPJPE (mm) }}}&
 \parbox[t]{2mm}{{\rotatebox[origin=c]{90}{Overall - P. Img.}}}
 \parbox[t]{1mm}{{\rotatebox[origin=c]{90}{MSE (kPa$^2$) }}}& 
 \parbox[t]{2mm}{{\rotatebox[origin=c]{90}{Overall - v2vP}}}
 \parbox[t]{1mm}{{\rotatebox[origin=c]{90}{MSE (kPa$^2$) }}}& 
 \parbox[t]{2mm}{{\rotatebox[origin=c]{90}{Body mass}}}
 \parbox[t]{1mm}{{\rotatebox[origin=c]{90}{MAE (kg) }}}& 
 \parbox[t]{2mm}{{\rotatebox[origin=c]{90}{Body height}}}
 \parbox[t]{1mm}{{\rotatebox[origin=c]{90}{MAE (mm) }}}\\

\hline
\hline		
BPWnet, 108K$\dagger$ & & $\times$ & $\times$ & 72.84 & 1.215 & 2.879 & 7.42 & 44.35  \\
\hline		
BPWnet, 108K$\dagger$ & $\times$ & & $\times$ & \textbf{68.62} & 1.296 & 2.981 & 7.62 & 65.97 \\
\hline		
BPWnet, 108K$\dagger$ & $\times$ & $\times$ & & 69.36 & \hspace{-1mm}16.326 & \hspace{-1mm}32.783 & \textbf{5.64} & \textbf{39.45} \\
\hline		
BPWnet, 108K$\dagger$ & $\times$ & $\times$ & $\times$ & 69.36 & \textbf{1.184} & \textbf{2.849} & \textbf{5.64} & \textbf{39.45} \\
\hline
\cline{1-9}						

\end{tabular}}
\end{center}
\end{table}


\textbf{CAL succeeds at both scaling and locally calibrating pressure map features.} Recall that the purpose of CAL is to calibrate $\mathcal{P}^+$ both by scaling it and by adjusting local features to better resemble features in the real pressure image $\mathcal{P}$. We tested the ability to achieve each of these purposes by adding a body mass normalization component to the output, which effectively scales $\mathcal{P}^+$ to the correct pressure range. The estimated body mass from BetaNet was used for this and the results are shown in Table~\ref{tab:results_table_CAL}. Without CAL, mass normalization greatly improves the pressure inference, but not to the extent that CAL does. This indicates that CAL does more to improve the features in $\mathcal{P}^+$ than only scaling it. We also compared a network that both uses a CAL and normalizes by mass, and observed a slight dip in performance.


\textbf{Improvements to blanket simulation may improve performance.} 
When training using only synthetic data (with and without blanket occlusions), pose estimation is substantially better when testing on real depth images that do not have blanket occlusions than testing on those that do. See Table~\ref{tab:results_table_pose} for reference. The same holds true when training on mixed synthetic and real data. On the other hand, when training using only real data (with and without blanket occlusions), pose estimation accuracy is comparable when testing on real depth images with and without blanket occlusions. This seems to indicate that the quality of synthetic blanket occlusions could be improved.

We manually adjusted simulation parameters to achieve realistic blanket folding characteristics. Besides blanket stiffness, we found that other FleX parameters such as the number of simulation substeps had an impact on the cloth behavior. 
Optimizing the synthetic blanket parameters to make them behave more like real coverings may improve the quality of synthetic data and boost performance when testing on real depth images with blanket occlusions; for example, the methods from Runia \etal~\cite{runia2020cloth} might merit future exploration.


\begin{table}[t!]
\caption{\label{tab:results_table_CAL} CAL feature calibration test - evaluated on the 22 subject test set. Pressure error shown with same metrics as previous tables.}
\vspace{-6mm}
\begin{center}
\renewcommand{\arraystretch}{1.1}
\scalebox{0.89}{\begin{tabular} {c|c|c|c|c}
\hline
 Network &    
 \parbox[t]{2mm}{{\rotatebox[origin=c]{90}{Normalize by}}}
 \parbox[t]{1mm}{{\rotatebox[origin=c]{90}{body mass}}}&
 \parbox[t]{2mm}{{\rotatebox[origin=c]{90}{CAL feature}}}
 \parbox[t]{1mm}{{\rotatebox[origin=c]{90}{calibration}}}&
 \parbox[t]{2mm}{{\rotatebox[origin=c]{90}{Overall - P. Img.}}}
 \parbox[t]{1mm}{{\rotatebox[origin=c]{90}{MSE (kPa$^2$) }}}& 
 \parbox[t]{2mm}{{\rotatebox[origin=c]{90}{Overall - v2vP}}}
 \parbox[t]{1mm}{{\rotatebox[origin=c]{90}{MSE (kPa$^2$) }}}\\

\hline
\hline		
BPWnet, 108K$\dagger$ & & & \hspace{-1mm}16.326 & \hspace{-1mm}32.783  \\
\hline			
BPWnet, 108K$\dagger$ & $\times$ & & 1.393 & 3.167 \\
\hline
BPWnet, 108K$\dagger$ & $\times$ & $\times$ & 1.195 & 2.893 \\
\hline		
BPWnet, 108K$\dagger$ &  & $\times$ & \textbf{1.184} & \textbf{2.849} \\
\hline
\cline{1-5}						

\end{tabular}}
\end{center}
\vspace{-0.1cm}
\end{table}

\textbf{Body pressure loss may improve performance.} The loss functions in BPWnet and BPBnet include terms computed at many different locations to provide better supervision. A loss directly computed based on the error in the inferred body pressure $\mathcal{P}_b$ merits future investigation. Recent differentiable geometry tools may enable such a loss computation and improve performance.

\textbf{Released materials can support future work.} In addition to the SLP-3Dfits human body annotations and the \methodname SD synthetic dataset used to train our model, we also publicly release 3D synthetic mesh data for the resting human, mattress, pressure sensing mat, and blanket. This may be useful for future work in the area of geometric learning.

\section{Opportunities for Future Work}

\begin{figure}
\begin{center}
\centering
\includegraphics[width=8.8cm]{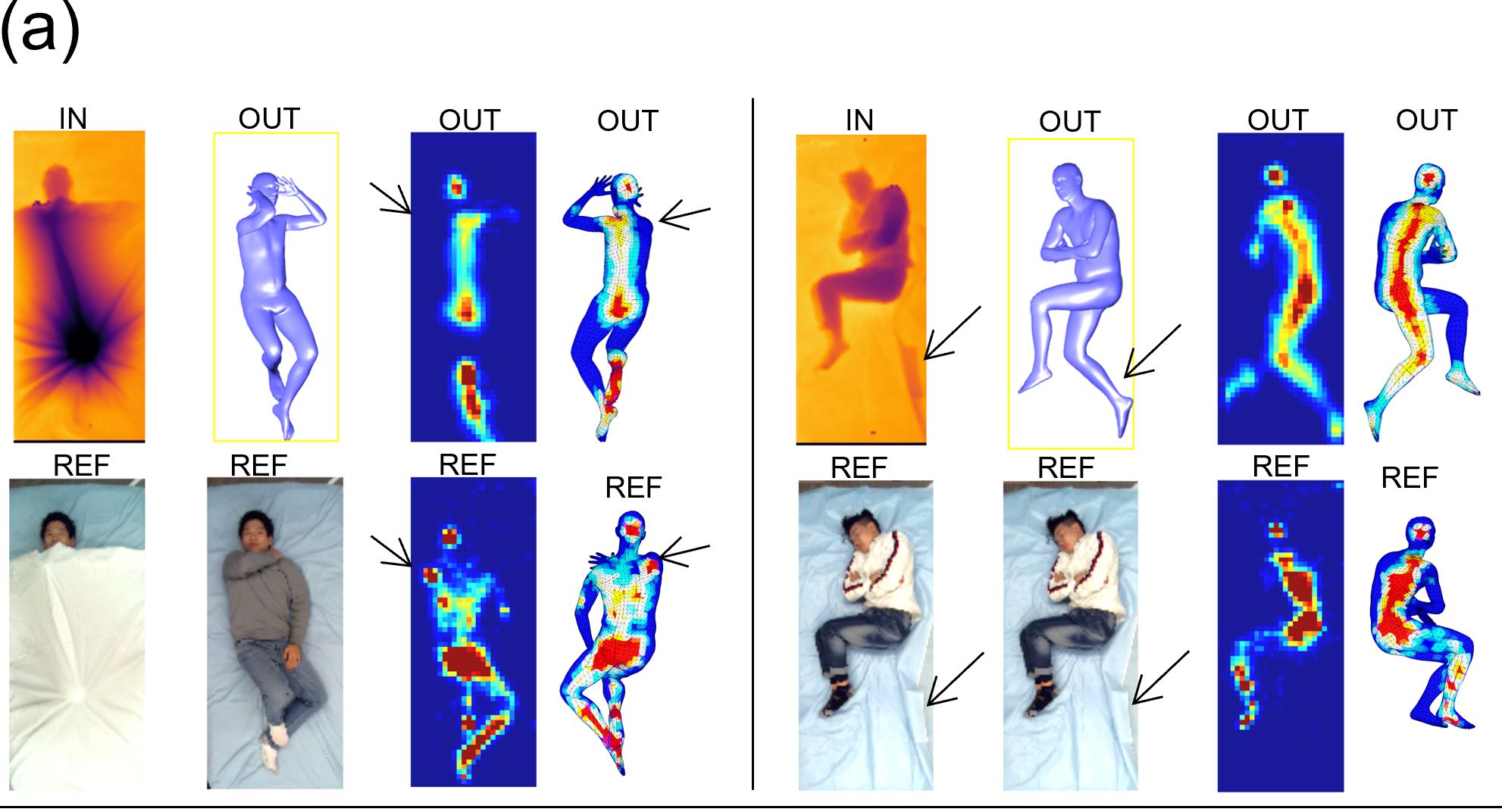}
\includegraphics[width=8.8cm]{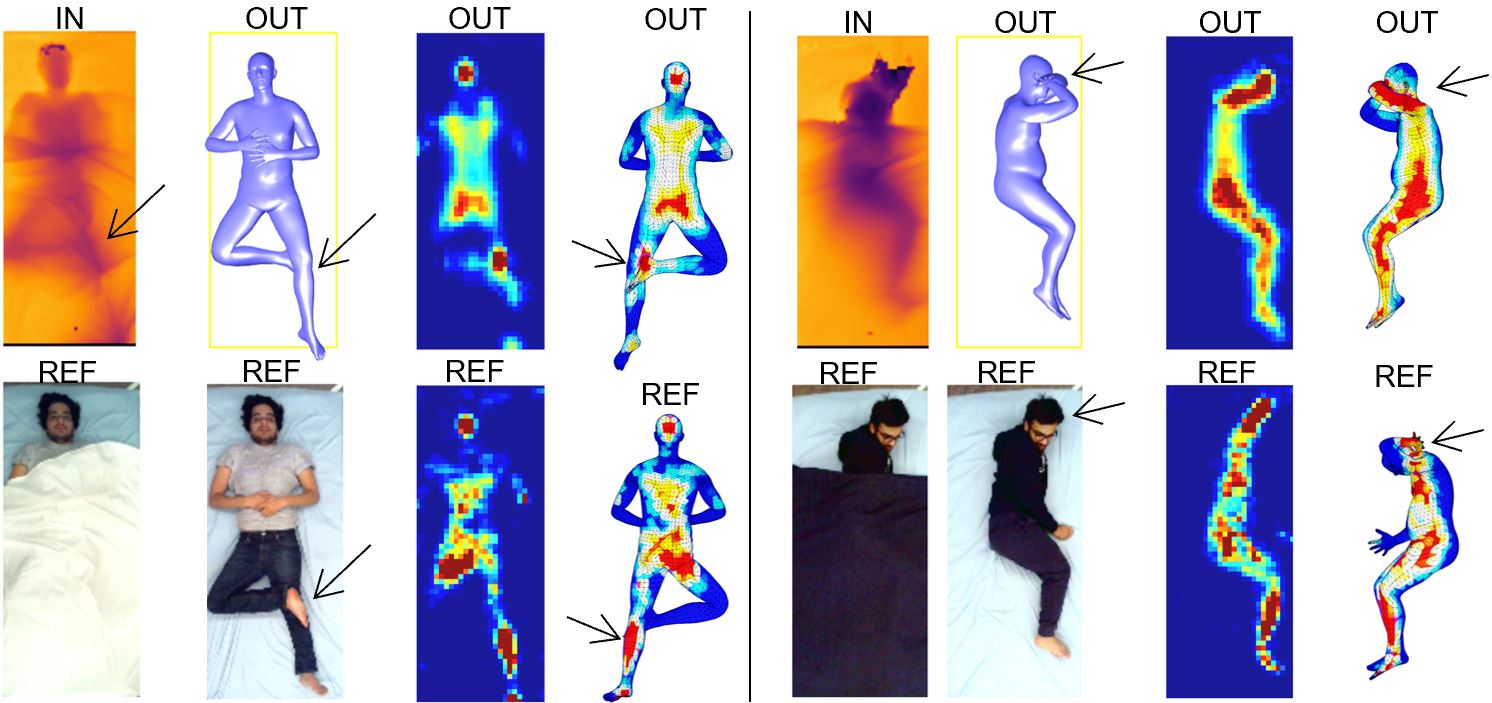}
\includegraphics[width=8.8cm]{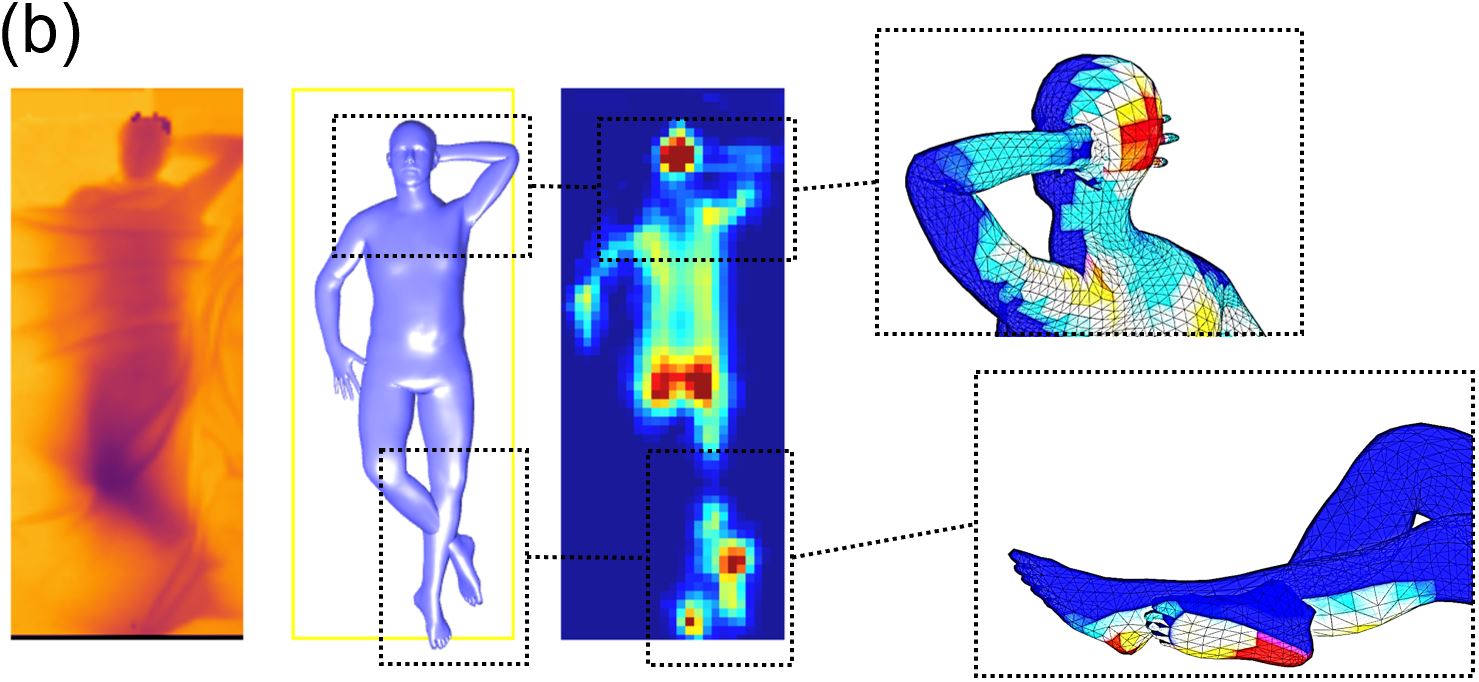}
\end{center}
\vspace{-0.5cm}
\caption{(a) Examples of errors when testing with BPWnet. (b) Limb interpenetration scenarios, also with BPWnet.}
\label{fig:failure_cases}
\end{figure}

While BPWnet exhibits promising performance and demonstrates the feasibility of using a depth sensor to infer body pressure, further research will be required to establish its clinical effectiveness. For example, the occurrence of false positives or false negatives may limit the system's ability to detect when a pressure injury is imminent. One example of a false negative is in the top left of case Fig.~\ref{fig:failure_cases} (a), where there is a peak pressure region on the person's right shoulder, but the system indicates there is no pressure on the right shoulder. The ability of the current network to generalize to clinical settings is also unclear. For example, pillows, nearby furniture, different types of bedding, medical instrumentation, and objects in bed, such as a mobile phone and other devices, would likely result in errors. 

Some types of errors might be difficult or impossible to overcome with a single frame from a depth camera. For example, in the top left error example in Fig.~\ref{fig:failure_cases} (a), the person's elbow and knee are elevated such that they push the blanket up into a tent like shape that reduces contact between the blanket and the person's body. This reduces depth information about the body surface and the system makes errors, including neglecting pressure on the hidden leg and one side of the body. 

For other types of errors, future work might achieve better performance. Fig.~\ref{fig:failure_cases} (a) shows examples of errors. On the top right, the sheet covering the bed folds upwards and the system mistakes it for the person's right leg. On the bottom left, the person crosses their right foot on top of their left knee and the system incorrectly estimates that the knee is on top of the foot. This leads the system to infer a peak pressure on the heel, rather than the calf. On the bottom right, the person assumes a pose that would be unusual when sleeping. The person rests their head on their left hand. The system misestimates the arm poses.  Additionally, the annotation method incorrectly labels the left hand as being behind the head rather than supporting it. 




Other errors relate to the body contacting itself. The network can output unnatural body part interpenetration. Fig.~\ref{fig:failure_cases}~(b) shows an example of this, where the left hand penetrates the head and the lower legs penetrate one another. Our network uses a fixed open hand pose, which may contribute to unnatural hand penetration errors. Self penetration of the 3D mesh body models does not occur frequently in the data because a mesh interpenetration term was used to create SLP-3Dfits, nor in the synthetic data because the physics simulations prevent it. Real human bodies have soft tissues that deform when in contact, which can be approximated as 3D model interpenetration, but the network outputs interpenetration that poorly matches soft tissue deformation. The problem is worsened by the frequent self-contact of limbs and body parts when a person rests. Our system also neglects pressure due to self contact and pressure differences due to the mass of one limb resting on another limb. Better accounting for the mechanics of self-contact might improve performance \cite{hassan2019resolving, muller2021self}. 



\section{Conclusion}
In summary, we presented a method to infer body pose and contact pressure from a depth image, which has the potential to automatically localize pressure injury risk areas using a consumer-grade depth camera. We described a method for annotating an existing human resting pose dataset with 3D body models, which we use for initializing a fast physics simulator and training and testing deep models. We generated a large synthetic resting pose dataset using physics simulations, which significantly boosts performance of our deep models. We introduced two deep learning models and compared their performance. The models were able to to accurately infer pose and contact pressure and outperform state-of-the-art methods for pose inference, even in the presence of visual occlusion from blankets.

\ifCLASSOPTIONcompsoc
  \section*{Acknowledgments}
\else
  \section*{Acknowledgment}
\fi

We thank Gerry Chen for his insightful feedback on the manuscript and Shuangjun Liu for assisting with the SLP dataset. This work was supported by the National Science Foundation Graduate Research Fellowship Program under Grant No. DGE-1148903, NSF award DGE-1545287, NSF award IIS-1514258, NSF award IIS-2024444, and AWS Cloud Credits for Research.

\section*{Disclosure} 
Charles C. Kemp owns equity in and works for Hello Robot, a company commercializing robotic assistance technologies. Henry M. Clever is entitled to royalties derived from Hello Robot's sale of products.

\ifCLASSOPTIONcaptionsoff
  \newpage
\fi



%
\bibliographystyle{IEEEtran}
%

\bibliography{main}

\newpage

%
\vspace{-0.5cm}
\begin{IEEEbiography}[{\includegraphics[width=1in,height=1.25in,clip, keepaspectratio]{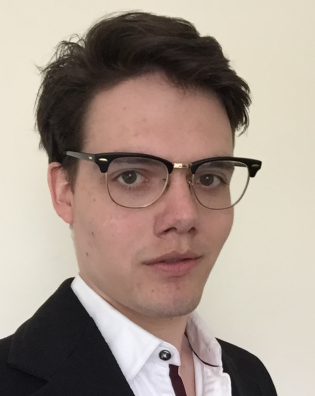}}]{Henry M. Clever}
received a B.S. in mechanical engineering from the University of Kansas, and a M.S. in mechanical engineering from New York University. Henry is currently pursuing a Ph.D. in robotics at the Georgia Institute of Technology in the Healthcare Robotics Lab. Henry's research interests include robot understanding in unstructured environments, haptic and vision perception of humans and robots, human-robot systems, physics simulation of humans and robots, and human pose estimation.
\vspace{-0.5cm}
\end{IEEEbiography}

\begin{IEEEbiography}
[{\includegraphics[width=1in,height=1.25in,clip, keepaspectratio]{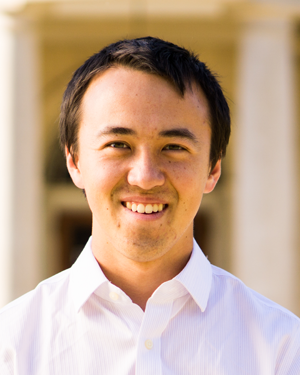}}]{Patrick L. Grady}
received a B.S. in computer science and electrical and computer engineering from Duke University. Patrick is currently pursuing a Ph.D. in robotics at the Georgia Institute of Technology in the Healthcare Robotics Lab. He is interested in hand-object interaction and computer vision for robots.

\vspace{-0.5cm}
\end{IEEEbiography}

\begin{IEEEbiography}[{\includegraphics[width=1in,height=1.25in,clip, keepaspectratio]{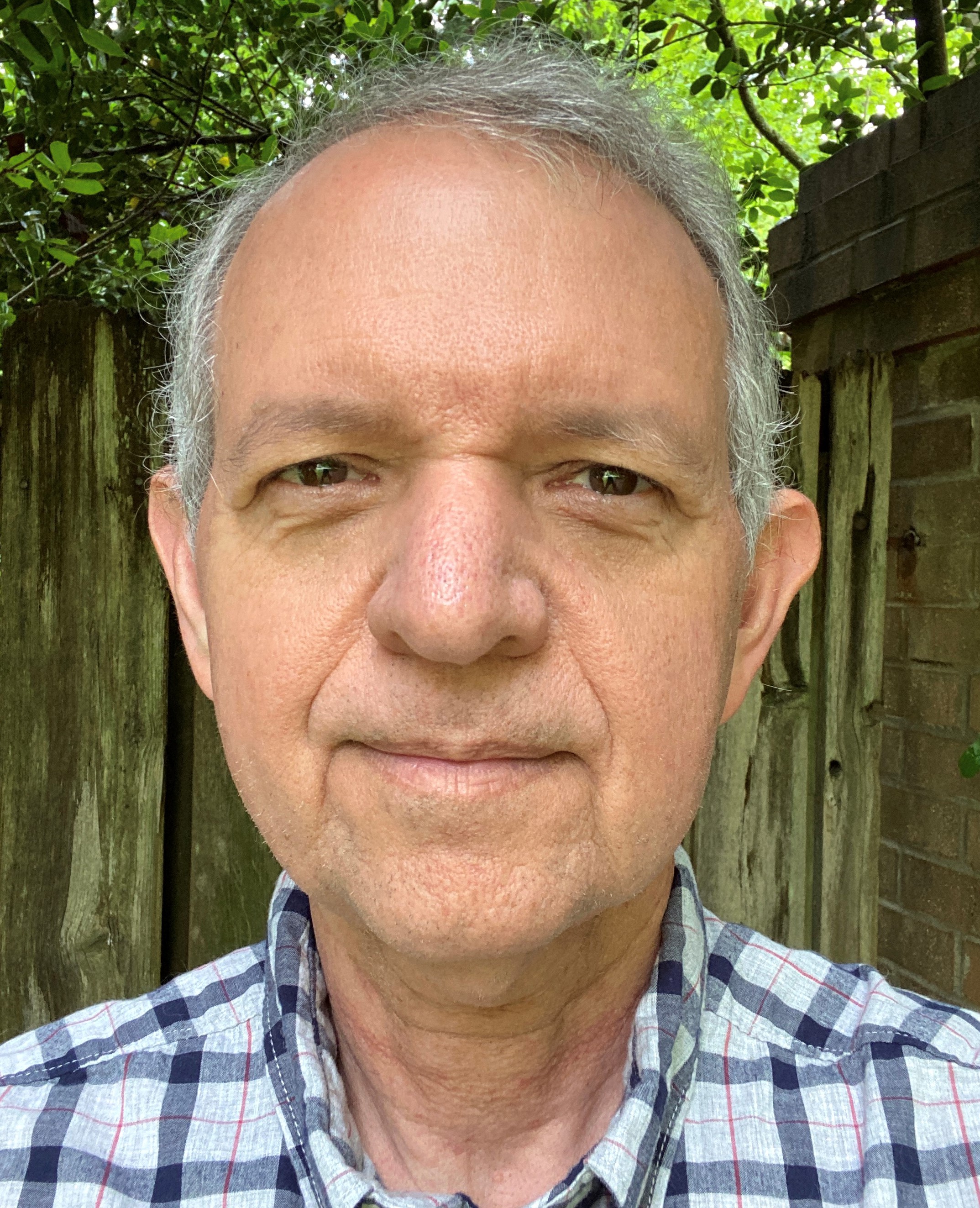}}]{Greg Turk} Greg Turk received a Ph.D. in computer science in 1992 from the University of North Carolina at Chapel Hill. He was a postdoctoral researcher at Stanford University for two years. He is currently a Professor at the Georgia Institute of Technology, where he is a member of the School of Interactive Computing and the Graphics, Visualization and Usability Center. His research interests include computer graphics, robotics, biological simulation and machine learning. He was the Technical Papers Chair for ACM SIGGRAPH 2008.  In 2012 he received the Computer Graphics Achievement Award from ACM SIGGRAPH for his computer graphics research. 
\vspace{-0.5cm}
\end{IEEEbiography}

\begin{IEEEbiography}[{\includegraphics[width=1in,keepaspectratio]{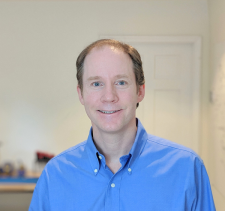}}]{Charles C. Kemp} (Charlie Kemp) is an Associate Professor at Georgia Tech in the Department of Biomedical Engineering with adjunct appointments in the School of Interactive Computing and the School of Electrical and Computer Engineering. In 2007, he founded the Healthcare Robotics Lab, which focuses on enabling robots to provide intelligent physical assistance in the context of healthcare. He earned a BS, an MEng, and a PhD from the Massachusetts Institute of Technology (MIT) in the areas of computer science and electrical engineering. 
\end{IEEEbiography}

\newpage 



\appendices

\section{Additional SLP-3Dfits results}\label{sec:app_3dfits}

\begin{figure}[h]
\begin{center}
\vspace{-4mm}
\includegraphics[width=8.8cm]{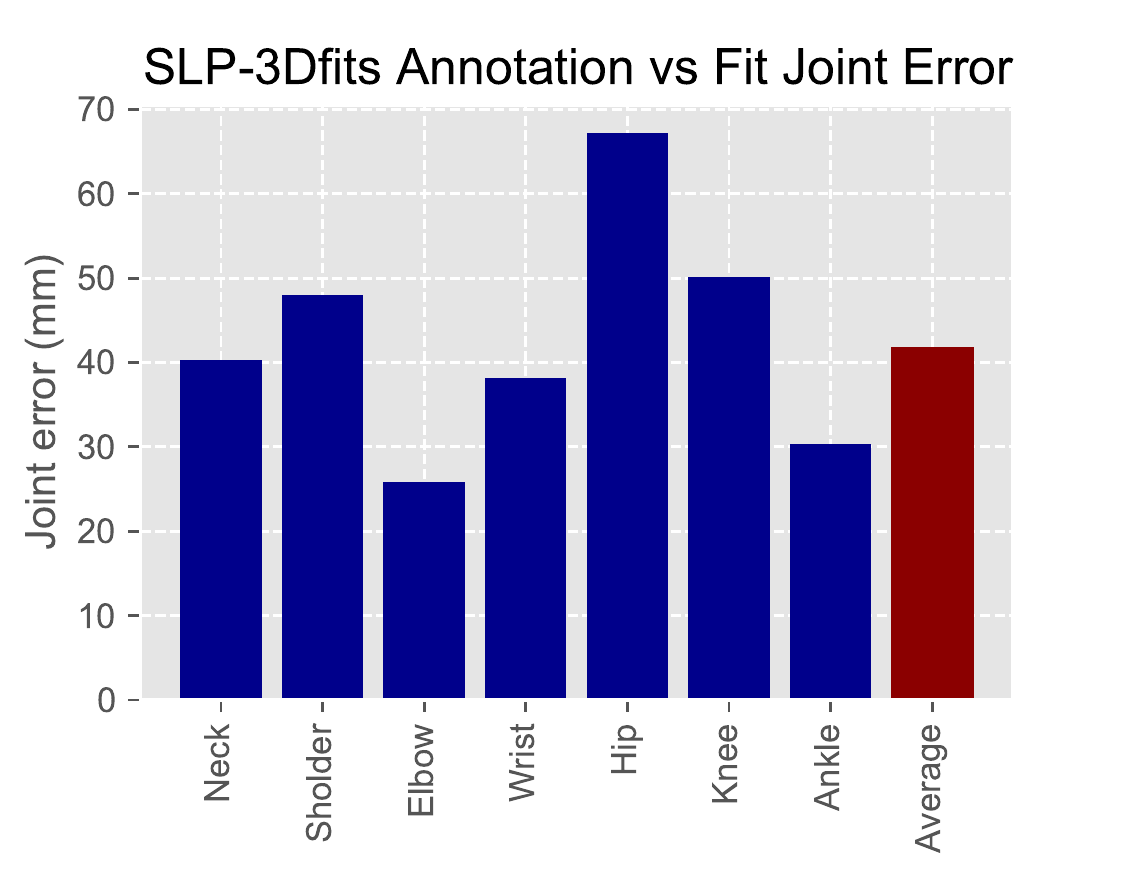}
\end{center}
\vspace{-0.7cm}
\caption{Per-joint error of SLP-3Dfits. The joint location of the SMPL model is compared to the 2D SLP annotation projected into 3D space.}
\label{fig:slp_fit_error}
\end{figure}

The per-joint error for $4545$ annotated poses in the SLP dataset is shown in Fig.~\ref{fig:slp_fit_error}. Note that the SLP dataset and SMPL model use different joint locations, causing high error for some joints.

\section{Weighting the capsulized human chain.}\label{sec:app_capsule_weight} We develop a method for assigning body part weights, based on Clever \etal~\cite{clever2020bodies}, for the physics simulation and synthetic dataset generation. 

A per-capsule weight for the articulated capsulized chain in DartFleX is computed based on the weight distribution for an average person and capsule volume ratios. First, we describe how capsule mass for the average person is assigned. We use average body mass and mass distribution values from Tozeren \cite{tozeren1999human}, and calculate capsule volumes from body shape. We assume the average human of gender $g\in \{M, F\}$ has a mass of $\bar{m_g}$, mass percentage distribution for body part $R$ of $\bar{X}_{R,g} \in \bar{\boldsymbol{X}}_{g}$, and SMPL body shape parameters $\bar{\boldsymbol{\beta}_g} = \mathbf{0}$. We define the mass of each capsule $c$ in an average person to be:

\begin{equation}
{\bar{m_c} = \bar{m_g} \bar{X}_{R,g} {\bar{\mathcal{V}}_{c,g} \over \bar{\mathcal{V}}_{R,g}}}
\end{equation}
where $\bar{\mathcal{V}}_{c,g}$ is the volume of capsule $c$ for a mean body shape $\bar{\boldsymbol{\beta}_g}$, and $ \bar{\mathcal{V}}_{R,g}$ is the sum of volumes for all capsules in body part $R$. Now, we describe how this capsule mass can be converted into masses for people of other shapes. To find the mass of some capsule $c$ for a body of particular shape $\boldsymbol{\beta}$, a capsule volume ratio between the particular person and an average person is used:

\begin{equation}{
m_c = \bar{m_c} {\mathcal{V}_c \over \bar{\mathcal{V}}_{c,g}}}
\end{equation}
where $\mathcal{V}_c$ is the volume of some arbitrary capsule. Note that when summing these capsules for the person, it substantially over-estimates the body density for heavy bodies and under-estimates for light bodies. Capsule mass is corrected by dividing by the total capsule volume ratio and multiplying by the SMPL mesh volume ratio. This latter provides a better mass-density ratio model.

\begin{equation}{
\widetilde{m}_c = m_c {\sum_{j=1}^N \bar{\mathcal{V}}_{j,g} \over \sum_{j=1}^N \mathcal{V}_j} \cdot {\mathcal{V}_{mesh, g} \over \bar{\mathcal{V}}_{mesh,g} }}
\end{equation}

where $N=20$ capsules, $\mathcal{V}_{mesh,g}$ is the SMPL mesh volume of a person with arbitrary shape in the home pose ($\boldsymbol{\Theta} = \mathbf{0}$), and $\bar{\mathcal{V}}_{mesh,g}$ is the SMPL mesh volume of a person with average shape ($\bar{\boldsymbol{\beta}_g} = \mathbf{0}$) in the home pose.

\begin{figure*}
\begin{center}
\centering
\includegraphics[width=18.2cm]{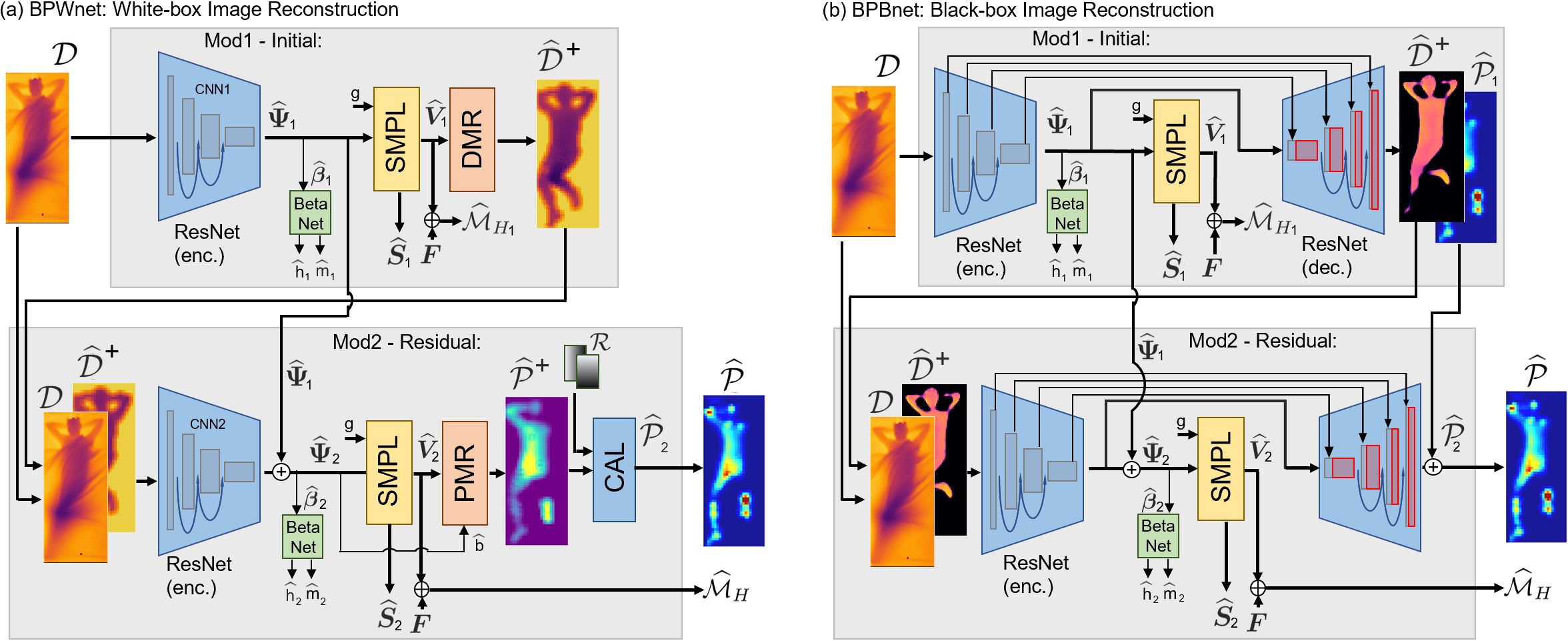}
\vspace{-9mm}
\end{center}
\caption{A comparison between BPWnet and its black-box variant, BPBnet. The BPWnet shown above is the same as Fig.~\ref{fig:networks} but contains more detail. BPBnet replaces the white-box DMR and PMR components with a learned ResNet34 decoder that shares weights with the encoder.}
\label{fig:networks_appendix}
\end{figure*}

\section{BPWnet Details}\label{sec:WBR_details}

This section provides explicit definitions for the loss function components introduced in Section~\ref{ssec:WBR}. Fig.~\ref{fig:networks_appendix}(a) shows BPWnet in more detail for reference to the loss variables. 

\textbf{Human pose loss computation.}
Recall that the encoder outputs estimated SMPL parameters $\boldsymbol{\widehat{\Psi}} = \big[ \boldsymbol{\hat{\beta}} \hspace{3mm} \boldsymbol{\widehat{\Theta}} \hspace{3mm} \boldsymbol{\hat{s}} \hspace{3mm} \boldsymbol{\hat{x}} \hspace{3mm} \boldsymbol{\hat{y}} \hspace{3mm}  \hat{b} \big]^{\top} \in \mathbb{R}^{89}$. The first two terms, $ \boldsymbol{\hat{\beta}} \in \mathbb{R}^{10}$ and $\boldsymbol{\widehat{\Theta}}\in \mathbb{R}^{69}$, contain the body shape and and joint angles, respectively, for the SMPL human model~\cite{loper2015smpl}. The terms $\boldsymbol{\hat{s}} \in \mathbb{R}^{3}$,  $\boldsymbol{\hat{x}} \in \mathbb{R}^{3}$, and  $\boldsymbol{\hat{y}} \in \mathbb{R}^{3}$ define the global transform of the SMPL model, with translation $\boldsymbol{\hat{s}}$ and continuous rotation parameters $\{x_{u}, x_{v}, x_{w}\} = \boldsymbol{\hat{x}}$, $\{y_{u}, y_{v}, y_{w}\} = \boldsymbol{\hat{y}}$ for 3 DOF, i.e. $\phi_{u}  = atan2(y_{u}, x_{u})$ and $\{\phi_{u}, \phi_{v}, \phi_{w} \} = \boldsymbol{\phi} \in \mathbb{R}^{3}$. The term $\hat{b} \in \mathbb{R}^{1}$ defines the distance between the camera and the bed. The encoder output $\boldsymbol{\widehat{\Psi}}$ is used to differentiably reconstruct a SMPL mesh with an embedded human kinematics model. As such, it also outputs 3D Cartesian joint positions $\boldsymbol{\widehat{S}} \in \mathbb{R}^{24 \times 3}$. The SMPL block outputs a set of vertices for the human, $\boldsymbol{\widehat{V}}_H \in \mathbb{R}^{6890 \times 3}$. The estimated human mesh $\mathcal{\widehat{M}}_H$ can be assembled from $\boldsymbol{\widehat{V}}_H$, as well as the faces $\boldsymbol{F}_H$, which are constant. 

Supervision is provided at multiple locations throughout the network to better utilize the fully observable nature of the synthetic data. Previous work has provided supervision on body shape~\cite{clever2020bodies} and on 3D Cartesian joint positions~\cite{kanazawa2018end}. We observed that adding global body rotation, SMPL joint angles, and camera-to-bed distance enables quicker and better learning at little computational cost, and define the following loss function:

\begin{equation}
\begin{split}
\mathcal{L}_{\textrm{SMPL}} :=  {1\over  N_{\beta} \sigma_\beta}\big|\big|\boldsymbol{\beta} - \boldsymbol{\hat{\beta}}\big|\big|^{}_1 + {1\over  N_{\Theta}\sigma_\Theta}\big|\big|\boldsymbol{\Theta} - \boldsymbol{\hat{\Theta}}\big|\big|^{}_1 \\
+ {1\over 6\sigma_{yx}}\Big(\big|\big|\boldsymbol{x} - \boldsymbol{\hat{x}}\big|\big|^{}_1 + \big|\big|\boldsymbol{y} - \boldsymbol{\hat{y}}\big|\big|^{}_1 \Big) \\
+ {1\over N_s\sigma_s}\sum_{j=1}^{N_s}\big|\big|\boldsymbol{s}_j - \boldsymbol{\hat{s}}_{j}\big|\big|^{}_2
+ {1\over \sigma_b}\big|\big|b - \hat{b}\big|\big|^{}_1
\end{split}
\label{eqn:smpl}
\end{equation}

where $N_{\beta}=10$ body shape parameters, $N_{\Theta} = 69$ joint angles, there are 6 parameters of continuous global rotation, $N_{s}=24$ Cartesian joint positions, and $\boldsymbol{s}_j \subset \boldsymbol{S}$ represents the Cartesian position of a single joint. Each term is normalized by standard deviations $\sigma_{\beta}$, $\sigma_{\Theta}$, $\sigma_{yx}$, $\sigma_s$, and $\sigma_{b}$, which are computed from the entire synthetic training dataset. Note that $\mathcal{L}_{\textrm{SMPL}}$ is sufficient to reconstruct the entire SMPL mesh, but does not require it during differentiation, so it can be computed efficiently. If the mesh is differentiably reconstructed, we may also define a loss on the SMPL mesh vertices:

\begin{equation}
\mathcal{L}_{\text{v2v}} := {1\over N_{V_H}\sigma^{}_{V_H}}\sum_{j=1}^{N_{V_H}}\big|\big|\boldsymbol{v}_j - \widehat{\boldsymbol{v}}_j\big|\big|_2
\label{eqn:v2v}
\end{equation}

where $\boldsymbol{v}_j \subset \boldsymbol{V}_H$ represents the Cartesian position of a single human mesh vertex, $N_{V_H} = 6890$ vertices, and the loss term is normalized by $\sigma^{}_{V_H}$.

\textbf{Image reconstruction loss computation.} Because the initial module is only necessary for producing a rough estimate and there are no learned parameters in DMR, no loss is computed on the depth maps $\widehat{\mathcal{D}}^+$. However, we do train the residual module with a loss on the reconstructed pressure maps $\widehat{\mathcal{P}}^+$. This minimizes the difference between the reconstructed pressure map for the estimated pose $\widehat{\mathcal{P}}^+$ and the ground truth pose $\mathcal{P}^+$. A binary contact map $\mathcal{\widehat{C}}_{\text{p}^+}$ is computed directly from $\mathcal{\widehat{P}}^+$, and we define the following loss function for PMR:

\begin{equation}
\mathcal{L}_{\mathcal{P}^+} := {1\over T\sigma^{}_{\mathcal{P}^+}}\big|\big|\mathcal{P}^+ - \widehat{\mathcal{P}}^+\big|\big|^2_2
+ {1\over T\sigma^{}_{\mathcal{C}_{\mathrm{p}^+}}}\big|\big|\mathcal{C}_{\mathrm{p}^+} - \widehat{\mathcal{C}}_{\mathrm{p}^+}\big|\big|^{}_1
\label{eqn:LQminus}
\end{equation}

where $T = 1728$ pressure map taxels and standard deviations $\sigma_\mathcal{Q^-}$ and  $\sigma_{\mathcal{C}_{q^-}}$ are similarly computed over the entire training dataset. Similarly, to train the feature calibration network CAL, we define a loss between the estimated and ground truth reconstructed pressure images $\mathcal{P}$ and their contact maps $\mathcal{C}_{\mathrm{p}}$:

\begin{equation}
\mathcal{L}_{\mathcal{P}} := {1\over T\sigma^{}_{\mathcal{P}}}\big|\big|\mathcal{P} - \widehat{\mathcal{P}}\big|\big|^{}_1
+ {1\over T\sigma^{}_{\mathcal{C}_{\mathrm{p}}}}\big|\big|\mathcal{C}_{\mathrm{p}} - \widehat{\mathcal{C}}_{\mathrm{p}}\big|\big|^{}_1
\label{eqn:LP}
\end{equation}

with standard deviations $\sigma_\mathcal{P}$ and $\sigma_{\mathcal{C}_p}$.

\section{Black-box Reconstruction Net (BPBnet)}\label{sec:resusmpl}

Here we describe BPBnet, a deep network with a black-box model of image reconstruction, shown in Fig.~\ref{fig:networks_appendix} (b). BPBnet uses a traditional black-box CNN for encoding depth images, as well as a black-box CNN for reconstructing depth and pressure images. Like BPWnet, it also produces an initial estimate with the first module, and uses the second module, to refine it through residual error~\cite{oberweger2015training, carreira2016human, clever2020bodies}. 

Each module has an encoder and decoder which extract features into a latent space and upsamples them back into images in a reverse fashion. While the middle latent space in such a network is often left unconstrained, it may encode specific targets, such as class labels~\cite{harouni2018universal}. Here it is used to encode the SMPL model parameters $\boldsymbol{\widehat{\Psi}}$, which are a sufficient representation to decode a fixed-frame depth image of the human body. For BPBnet, $\boldsymbol{\widehat{\Psi}} \in \mathbb{R}^{88}$, since it drops the camera to bed distance term present in BPWnet. $\mathcal{L}_{\textrm{SMPL}}$ for BPBnet similarly drops the last term on Eq.~\ref{eqn:smpl}. The black-box ResNet decoder outputs occlusion-free depth images, pressure images, and their associated binary contact maps.

\textbf{Black-box decoding.} Vector $\boldsymbol{\widehat{\Psi}}$ is additionally fed into a decoder, which uses learnable weights to reconstruct an occlusion-free depth image of the human body, $\mathcal{\widehat{D}}^{+}$, and a pressure image $\mathcal{\widehat{P}}$. BPBnet employs a U-Net~\cite{ronneberger2015u} style architecture, where the encoder and decoder share feature maps at each stage of convolution. We define a loss between the estimated and ground truth reconstructed depth images $\mathcal{D}^+$, as well as the contact maps $\mathcal{C}_{d^+}$, which are a binary function of $\mathcal{D}^+$:

\begin{equation}
\mathcal{L}_{\mathcal{D}^+} := {1\over T\sigma^{}_{\mathcal{D}^+}}\big|\big|\mathcal{D}^+ - \widehat{\mathcal{D}}^+\big|\big|^2_2
+ {1\over T\sigma^{}_{\mathcal{C}_{\mathrm{d}^+}}}\big|\big|\mathcal{C}_{\mathrm{d}^+} - \widehat{\mathcal{C}}_{\mathrm{d}^+}\big|\big|^{}_1
\label{eqn:LQplus}
\end{equation}

\textbf{BPBnet training strategy.} We train Mod1 and Mod2 separately, and use the same pre-trained BetaNet from Section~\ref{ssec:betanet} with locked weights for both modules. We first train Mod1, with the following loss function:

\begin{equation}
\mathbb{L}_{\textrm{BPB}_1} = \mathcal{L}_{\textrm{BetaNet}_1}+\mathcal{L}_{\textrm{SMPL}_1}+\mathcal{L}_{\mathcal{D}^{+}_1} +\mathcal{L}_{\mathcal{P}_1} 
\label{eqn:BB1}
\end{equation}

where subscripts $1$ on the terms indicate a loss computed from Mod1 estimates. This differs from the loss in Eq.~\ref{eqn:WB1} in that it contains image reconstruction terms for training the decoder. Next, we precompute a set of reconstructed depth and contact maps $\{\boldsymbol{\mathcal{\widehat{D}}}^+,  \boldsymbol{\widehat{\mathcal{C}}}_{\mathrm{d}^+}\}$ by pushing the entire depth image dataset $\boldsymbol{\mathcal{D}}$ through Mod1. We train Mod2 with a dataset consisting of inputs $\{\boldsymbol{\mathcal{D}}, \boldsymbol{\mathcal{\widehat{D}}}^+, \boldsymbol{\widehat{\mathcal{C}}}_{\mathrm{d}^+}\}$. Mod2 learns a spatial residual correction to improve estimates $\boldsymbol{\widehat{\Psi}}_1$ and $\mathcal{\widehat{P}}_1$, which are shown on the bottom of Figure~\ref{fig:networks_appendix} (a). As such, the Mod2 encoder outputs a correction $(\boldsymbol{\widehat{\Psi}}_2 - \boldsymbol{\widehat{\Psi}}_1)$ and the decoder outputs $(\mathcal{\widehat{P}}_2 - \mathcal{\widehat{P}}_1)$. The following loss is computed to train Mod2 in BPBnet:

\begin{equation}
\mathbb{L}_{\textrm{BPB}_2} = \mathcal{L}_{\textrm{BetaNet}_2}+\mathcal{L}_{\textrm{SMPL}_2}+\mathcal{L}_{\text{v2v}_2} +\mathcal{L}_{\mathcal{P}_2} 
\label{eqn:BB2}
\end{equation}

In contrast to Mod1 in BPWnet, the depth reconstruction term is omitted and a term is added for the SMPL mesh vertices, which slows training but refines estimation quality. The Mod2 encoder in BPBnet outputs $\widehat{\mathcal{M}}_{H,2}$ and the decoder outputs $\widehat{\mathcal{P}}_2$.

\section{Normalizing Pressure by Body Mass}\label{sec:massnorm}
In theory, the sum of pressure image values times the area should equal the weight of a person. However, in practice, seemingly innocuous changes such as placing an extra blanket between an object and the pressure mat can alter recorded data on some mats~\cite{clever2020bodies}. We normalize all ground truth pressure images in the SLP dataset and in the synthetic dataset by the body mass of each person, using the following equation:


\begin{equation}
\mathcal{P} = \mathcal{P}' {mg \over \sum p_{xy}'A_t}
\label{eqn:mn}
\end{equation}

where $m$ is the body mass, $g$ is gravitational acceleration, $p_{xy}'$ is the pressure measured by a single taxel (tactile pixel) in $\mathcal{P}'$, and $A_t$ is the surface area of a single taxel on the pressure mat. We assume the blanket has negligible mass. As such, normalized ground truth pressure $\mathcal{P}$ projects onto the human mesh $\mathcal{M}_{H}$ following Eq.~\ref{eqn:proj_M}.

\section{Co-registering Real and Synthetic Images}\label{sec:appendix_coreg}
We position the simulated pinhole depth camera to match the camera in the SLP dataset using the given camera intrinsics. In the SLP dataset, the various visual modalities (depth, RGB, IR) are aligned to the pressure mat via a calibration process, such that the 3D world origin in point cloud space is located at the top left corner of the pressure mat. However, the twin bed mattress and pressure mat used in the SLP dataset are of a different resolution and size than those in the simulation. Using manufacturer specifications and real measurements, the real SLP pressure images are converted to a lower resolution and slightly smaller size so they match the synthetic images. We perform similar conversions of depth imagery. It is ambiguous how far the corner of the pressure mat is from the corner of the bed. Our simulated bed and pressure mat are based on precise measurements from a real bed and mat in our previous work~\cite{clever2020bodies}. While both environments are square and the rotational discrepancy is negligible, we precisely align the reference frame origin translation between the real SLP dataset and our synthetic dataset. We use a grid search to do this. 


\section{Blanket Configuration Partition Details}\label{sec:app_blanket_partition}

The two blanket partitions are represented by translations $\boldsymbol{s}_B^{\ast}$ and $\boldsymbol{s}_B^{\ast\ast}$. In the first, we set the initial blanket position to:

\begin{equation}
\boldsymbol{s}_B^{\ast} = 
\left \{   
  \begin{tabular}{c}
${1\over 2} s_{M,1}, \hspace{2mm} s_{neck,2} - {1\over 2} s_{M,2}, \hspace{2mm}\xi_B $
  \end{tabular}
\right \}
\end{equation}

where $\boldsymbol{s}_{M}$ contains the mattress dimensions, $s_{neck,2}$ is the distance to the person's neckline from the origin, and the translation of the blanket is measured from its center and the world reference frame is located at the bottom left corner of the mattress. The constant $\xi_{B}$ is a distance above the resting human body, where $\xi_{B}$ is always above the highest joint position. In the second partition, we randomly sample across the person in bed, using:

\begin{equation}
\boldsymbol{s}_B^{\ast\ast}\hspace{-1mm} = \hspace{-1mm}\left \{ \hspace{-3mm}  
  \begin{tabular}{c}
  \vspace{1mm}
  $ \mathcal{U}(-s_{j,1,min},  s_{j,1,max})$\\
  \vspace{1mm}
  $s_{\alpha} \hspace{-1mm} + \hspace{-1mm}\mathcal{U} \big(-{s_{neck,2}-s_{j,2,min}\over 2 },  {s_{neck,2}-s_{j,2,min}\over 2 }\big)$\\
  $\xi_B $
  \end{tabular}
\hspace{-3mm}\right \}^{\hspace{-1mm}\top}
\end{equation}

where $s_{\alpha} =  s_{neck,2} - {1\over 2} s_{M,2} - 0.4$, so the distribution of the longitudinal distance $s_{B,2}^{\ast\ast}$ is centered around a location $0.4$ meters below the neckline. We shift down by $0.4$ so that the bottom edge of the blanket does not leave the legs too uncovered, which is uncommon in the SLP dataset. The range of the distribution is set to be equal to the longitudinal distance between the neck, $s_{neck,2}$, and the joint furthest toward the bottom of the bed, $s_{j,2,min}$ (typically a foot). The uniform distribution of the lateral shift $s_{B,1}^{\ast\ast}$ across the surface of the bed is set to a range between the furthest extend of a human joint in either direction (i.e. $-s_{j,1,min},  s_{j,1,max}$). 







\section{Network Computation Performance}\label{sec:computation}
BPWnet method has two computational advantages. First, it is a more parsimonious model. Using white-box image reconstruction with DMR and PMR in Fig.~\ref{fig:networks_appendix} instead of the learned ResNet decoder BPBnet reduces the learnable network parameters by 12\%. Second, it is more memory efficient. Our BPBnet implementation uses almost 23 GB of GPU RAM, which will not fit on many consumer grade GPUs. The white-box model can be tuned to sacrifice speed for improved memory while keeping an encoder training batch size of 128 -- at a cost of doubling the training time, the memory footprint reduces from 17 GB to 11.5 GB.




\end{document}